\documentclass[lettersize,journal]{IEEEtran}
\usepackage{amsmath,amssymb,amsfonts,amsthm}
\usepackage{algorithmic}
\usepackage{algorithm}
\usepackage{array}
\usepackage{arydshln}
\usepackage[caption=false,font=normalsize,labelfont=sf,textfont=sf]{subfig}
\usepackage{textcomp}
\usepackage{stfloats}
\usepackage{url}
\usepackage{color}
\usepackage{verbatim}
\usepackage{graphicx}
\usepackage{cite}
\newtheorem{definition}{{Definition}}
\usepackage{hyperref}
\usepackage[normalem]{ulem}
\usepackage{multirow}
\usepackage{enumitem}
\usepackage{float}
\usepackage{caption}
\usepackage{subcaption}
\usepackage{makecell}

\usepackage{xspace}
\newcommand{\etal}{et al.\xspace}

\newcommand{\ie}{i.e.,\xspace}

\newcommand{\graph}{\emph{HEIG}\xspace}
\newcommand{\model}{\emph{Meta-IFD}\xspace}

\begin{document}

%
\title{Enhancing Ethereum Fraud Detection via Generative and Contrastive Self-supervision}


        
        

\author{Chengxiang Jin, Jiajun Zhou, Chenxuan Xie, Shanqing Yu, Qi Xuan,~\IEEEmembership{Senior Member,~IEEE}, Xiaoniu Yang

\IEEEcompsocitemizethanks{
\IEEEcompsocthanksitem This work was supported by the China Postdoctoral Science Foundation under Grant Number 2024M762912, by the Postdoctoral Science Preferential Funding of Zhejiang Province of China under Grant ZJ2024060, by the Key R\&D Program of Zhejiang under Grant 2022C01018, by the National Natural Science Foundation of China under Grant U21B2001. (Corresponding author: Jiajun Zhou.)
\IEEEcompsocthanksitem C. Jin, C. Xie, S. Yu and Q. Xuan are with the Institute of Cyberspace Security, College of Information Engineering, Zhejiang University of Technology, Hangzhou 310023, China, with the Binjiang Institute of Artificial Intelligence, ZJUT, Hangzhou 310056, China. E-mail: jincxiang@zjut.edu.cn.
\IEEEcompsocthanksitem J. Zhou are with the Institute of Cyberspace Security, College of Computer Science and Technology, Zhejiang University of Technology, Hangzhou 310023, China, with the Binjiang Institute of Artificial Intelligence, ZJUT, Hangzhou 310056, China. E-mail: jjzhou@zjut.edu.cn.
\IEEEcompsocthanksitem X. Yang is with the Institute of Cyberspace Security, Zhejiang University of Technology, Hangzhou 310023, China, and also with Science and Technology on Communication Information Security Control Laboratory, Jiaxing 314033, China. Email: yxn2117@126.com.
}}



\maketitle

\begin{abstract}
The rampant fraudulent activities on Ethereum hinder the healthy development of the blockchain ecosystem, necessitating the reinforcement of regulations. However, multiple imbalances involving account interaction frequencies and interaction types in the Ethereum transaction environment pose significant challenges to data mining-based fraud detection research. To address this, we first propose the concept of meta-interactions to refine interaction behaviors in Ethereum, and based on this, we present a dual self-supervision enhanced Ethereum fraud detection framework, named \model. This framework initially introduces a generative self-supervision mechanism to augment the interaction features of accounts, followed by a contrastive self-supervision mechanism to differentiate various behavior patterns, and ultimately characterizes the behavioral representations of accounts and mines potential fraud risks through multi-view interaction feature learning. Extensive experiments on real Ethereum datasets demonstrate the effectiveness and superiority of our framework in detecting common Ethereum fraud behaviors such as Ponzi schemes and phishing scams. Additionally, the generative module can effectively alleviate the interaction distribution imbalance in Ethereum data, while the contrastive module significantly enhances the framework's ability to distinguish different behavior patterns. The source code will be available in \url{https://github.com/GISec-Team/Meta-IFD}.

\end{abstract}


\section{Introduction}
Ethereum~\cite{wood2014ethereum}, an open-source blockchain platform, is rapidly gaining prominence due to its revolutionary smart contract technology. 
Through smart contracts~\cite{zheng2020overview}, developers can establish protocols that autonomously execute, manage, and enforce contractual terms without the involvement of any third-party intermediary.
This novel paradigm has given rise to the concept of decentralized finance (DeFi)~\cite{zetzsche2020decentralized} in the financial domain. In Ethereum, DeFi applications are implemented via smart contracts, allowing users to directly interact with financial agreements for activities such as lending, trading, investing, and insurance. The automation and transparency of smart contract execution provide users with a more liberated financial experience, enabling global access to financial services without geographical or institutional constraints.

However, as the deployment of Ethereum in the financial domain surges, its technological characteristics raise security concerns for the financial ecosystem. First, decentralization means there is no central authority for oversight and regulation, allowing fraudsters to operate with impunity. Second, anonymity enables malicious actors to hide their real identities, exacerbating the challenges associated with crime detection. Third, the immutability of contracts makes it difficult to fix code vulnerabilities once they are discovered.
Furthermore, the global liquidity of cryptocurrencies poses challenges to the coordination of regulatory frameworks across diverse jurisdictions.
Despite the increasing focus on cryptocurrency finance security in recent years, a multitude of fraudulent activities persist, resulting in substantial financial losses.
According to the Bangkok Post\footnote[1]{\href{https://www.bangkokpost.com/thailand/general/2644790/cops-arrest-5-foreigners-over-links-to-b2-7bn-scam}{https://www.bangkokpost.com/}}, the Cyber Crime Investigation Bureau (CCIB) arrested individuals involved in a fraudulent virtual currency investment platform that swindled more than \$27 million. 
Besides, TRM Labs\footnote[2]{\href{https://www.trmlabs.com/post/inebreaknorth-korean-hackers-stole-600-million-in-crypto-in-2023}{https://www.trmlabs.com/}}  reported that North Korean hackers stole at least \$600 million in cryptocurrency in 2023. 
Therefore, the regulation of Ethereum cryptocurrency transactions is imperative to ensure financial security.

\begin{figure}[htb]
  \centering
  \includegraphics[width=\linewidth]{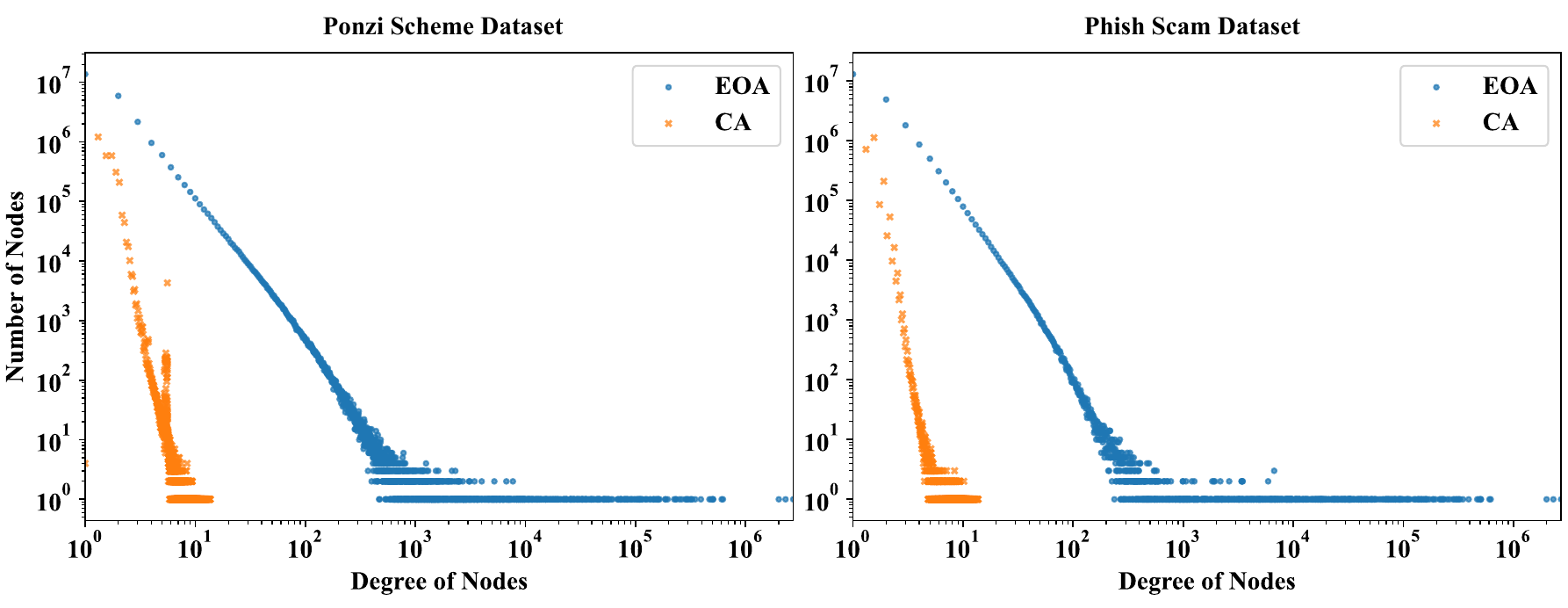}
  \caption{Illustration of the imbalanced distribution in account interaction frequency. Various types of accounts display long-tailed distributions for both types of fraud.}
  \label{fig: degree-dis}
\end{figure}

From the perspective of data mining, existing Ethereum fraud detection studies mainly employ machine learning (ML)~\cite{aziz2022lgbm} and graph modeling analysis~\cite{zhou2022behavior,li2022phishing} to uncover potential fraudulent risks in transaction records and smart contracts. ML-based methods mainly extract manual features from Ethereum data and train ML models to identify and predict potential fraudulent behaviors. However, such methods heavily rely on expert knowledge for designing manual features, limiting their generalizability to diverse blockchain scenarios. On the other hand, graph-based methods mainly use Ethereum data to construct transaction graphs and design graph learning models to identify malicious accounts and fraudulent activities. Although such methods can effectively capture account behavior patterns and improve the effectiveness and interpretability of fraud detection, they rely on constructing transaction graphs and designing graph learning models. Moreover, both studies overlook certain challenges inherent in Ethereum transaction scenarios to some extent:
1) \textbf{Frequent and complex transaction activities}. Ethereum continuously generates a large number of transaction records, involving different types of accounts and interaction patterns, which imposes higher requirements on the models' representation capabilities;
2) \textbf{Imbalanced interaction frequency distribution}. Ethereum accounts exhibit varying levels of interaction activity, as shown in Fig.~\ref{fig: degree-dis}, and such imbalance in interaction frequency distribution can prevent the models from effectively learning the behavior patterns of low-activity accounts;
3) \textbf{Imbalanced account behavior distribution}. Accounts of the same type on Ethereum may participate in diverse transaction activities, displaying distinct behavior preferences. This imbalance in behavior distribution can lead to biases in the models' characterization of account features.

To overcome the limitations of existing studies and address the challenges in Ethereum, we propose a dual self-supervision enhanced Ethereum fraud detection framework, named \model. Specifically, this framework first defines the concept of meta-interactions to describe fine-grained Ethereum interaction behaviors and introduces a generative self-supervised mechanism to design a fine-grained Ethereum interaction feature generation module. This module can augment multi-view interaction features for accounts with low activity levels and low-frequency interaction types, alleviating the imbalance in behavior distribution. Secondly, the framework introduces a contrastive self-supervised mechanism to design a coarse-grained Ethereum interaction feature contrast module, which can improve the framework's ability to distinguish different account behavior patterns. Finally, the framework designs a multi-view Ethereum interaction feature learning module, which effectively characterizes account features and captures potential fraudulent behaviors through account type-specific feature encoding, aggregation, and propagation processes.
We conduct extensive evaluation and comparative experiments in real Ethereum transaction scenarios to validate the effectiveness and superiority of our proposed framework in detecting common Ethereum fraud behaviors, such as Ponzi schemes and phishing scams.
The main contributions of this paper can be summarized as follows:
\begin{itemize}[leftmargin=10pt]
  \item We define the concept of meta-interactions to describe fine-grained Ethereum interaction behaviors.
  \item We design a dual self-supervision enhanced Ethereum fraud detection framework. By incorporating fine-grained interaction feature generation, coarse-grained interaction feature contrast, and multi-view feature learning processes, our framework alleviates the imbalance in account behavior distribution, enhances the ability to distinguish different behavior patterns, and ultimately improves the effectiveness of Ethereum fraud detection.
  \item Extensive experiments on real Ethereum data demonstrate the effectiveness and superiority of our framework in detecting Ethereum fraud behaviors.
\end{itemize}

\section{Related Work} \label{sec:related work}
This section reviews research on fraud detection in Ethereum, primarily focusing on the detection of Ponzi schemes and phishing scams. The related studies can be broadly categorized into two types: machine learning-based approaches and graph analysis-based approaches.

\subsection{Machine Learning-based Fraud Detection}
This type of methods typically utilizes manual feature engineering to extract predefined features from Ethereum data and train machine learning models to identify and predict potential fraudulent behaviors, such as Ponzi schemes and phishing scams.
Since such methods design targeted manual features based on different types of fraud, this subsection discusses related studies for different fraud separately.

\subsubsection{Ponzi Scheme Detection}
Ponzi schemes~\cite{moore2012postmodern} in Ethereum involve writing and deploying smart contracts that promise high returns to attract investors to transfer funds to the contract address. The returns for early investors come from the funds of subsequent investors rather than legitimate investment profits. 
Due to the anonymity of Ethereum, fraudsters can easily hide their identities and transfer funds, increasing the difficulty of regulation and tracking. When there are not enough new investors, the funds in the contract are insufficient to pay the high returns, leading to the collapse of Ponzi schemes.

Since Ponzi schemes on Ethereum are typically deployed on contract accounts, analyzing smart contract code for Ponzi scheme detection is a straightforward option.
Bartoletti \etal~\cite{bartoletti2020dissecting} analyze the source codes of Ponzi contract and categorize Ponzi schemes into four types based on their behavioral patterns. 
Onu \etal~\cite{onu2023detection} compare the performance of random forest, neural network, and k-nearest neighbors algorithms in Ponzi scheme detection, concluding that the random forest model achieves the best detection performance with the fewest features, enabling early detection of Ponzi schemes. 
Zheng \etal~\cite{zheng2023securing} extract contract account features from multiple views, including bytecode, semantics, and developer information, and develop a multi-view cascaded ensemble model to detect Ponzi schemes in newly created contract accounts. 
However, only the bytecode of smart contracts is publicly available, while the source code is not always. As a result, detection methods based on code analysis might not fully identify Ponzi schemes concealed within smart contracts.
Some studies combine more accessible transaction records to detect Ponzi schemes by analyzing the behavior of contract accounts. 
Hu \etal~\cite{hu2021transaction} summarize four types of fraudulent behavior patterns by analyzing account transaction records, and then construct 14-dimensional transaction features and train LSTM~\cite{chen2019lstm} model for fraud detection. 
Chen \etal~\cite{chen2018detecting} extract opcode frequency features from contract code and 13 transaction features from transaction records, combining these features to train XGBoost model for Ponzi scheme detection. 
In addition to these two types of features, Zhang \etal~\cite{zhang2021detecting} further integrate bytecode features and employ LightGBM for more efficient fraud detection. 
Considering the dynamic nature of transactions, Wang \etal~\cite{wang2023temporal} construct temporal transaction features based on timestamps and combine them with contract code features for early detection of Ponzi schemes.

\subsubsection{Phish Scam Detection}
Phishing scams on Ethereum~\cite{chen2020phishing} employ deceptive tactics to lure victims into disclosing their private keys or transferring funds to accounts controlled by attackers. Since phishing scams typically do not rely on contract functionality, detection against phishing scams usually focuses only on transaction records.
Chen \etal~\cite{chen2020phishing} extract relevant features of the target accounts and their interaction objects from transaction records and utilize LightGBM for phishing detection.
Kabla \etal~\cite{kabla2022eth} utilize a voting technique based on multi-ranking methods to select important features for phishing detection, which improves detection efficiency while reducing feature dimensions.
Hu \etal~\cite{hu2023bert4eth,hu2024zipzap} first serialize transaction features according to timestamps and then train BERT for phishing detection.
Phishers often disguise themselves by frequently interacting with normal accounts, thereby disrupting fraud detection. Several studies extract account features from multiple views to break through the disguise.
Ghosh \etal~\cite{ghosh2023investigating} extract relevant features from the temporal and network views in addition to the account's intrinsic features, fuse the multi-view features and feed them into various classifiers for phishing detection.
Xu \etal~\cite{xu2024ewdps} propose a multi-layer defense system to effectively simulate and track the dynamic evolution of phishing scams, achieving early to mid-term phishing fraud warnings.

\begin{table}
  \centering
  \renewcommand\arraystretch{1.2}
  \caption{Notations and descriptions.}
  \label{tab: notation}
  \resizebox{\linewidth}{!}{
  \begin{tabular}{ll} 
  \hline\hline
  Notation                      & Description                           \\ 
  \hline
  $G$                     & Heterogeneous Ethereum interaction graph          \\
  $\mathcal{V},\mathcal{E}$       & Set of account nodes / interaction edges    \\
  $\mathcal{T}_\mathcal{V}$, $\mathcal{T}_\mathcal{E}$                        & Account / Interaction type mapping function    \\
  $\mathcal{R}$                & Set of meta-interactions                            \\
    $M$   & Number of the views \\
      $\boldsymbol{X}$        &  Account feature matrix      \\
      $\hat{\mathcal{X}}$        &  Set of augmented multi-view interaction features       \\
    $\hat{\mathcal{H}}$  & Set of type-specific multi-view interaction features  
      \\
     $\bar{\mathcal{H}}$   & Set of aggregated multi-view account features \\
    $\bar{\boldsymbol{H}}$  & Account feature matrix after multi-view fusing
      \\
      $\boldsymbol{H}$  & Final account representations \\
  \hline\hline
  \end{tabular}}
\end{table}

\subsection{Graph-based Fraud Detection}

This type of methods primarily utilize transaction records to construct large-scale transaction graphs and design graph-based approaches to extract account behavior features for fraud detection. Given the broad applicability of graph methods in analyzing diverse account behavior patterns, this subsection does not specifically discuss the detection of different frauds separately.
Wu \etal~\cite{wu2020phishers} guide a random walk process on the Ethereum transaction graph using transaction amounts and timestamps to learn the structural representation of accounts, ultimately using SVM classifier for phishing detection. 
Similarly, Li \etal~\cite{li2022internet} calculate the similarity between accounts by considering the structural homogeneity and transaction homogeneity of the transaction graph, guiding random walks to learn account representations.
Tharani \etal~\cite{tharani2024unified} integrate transaction features and attributes of the transaction graph, employing GraphSAGE~\cite{hamilton2017inductive} to acquire account representations, which are subsequently jointly fed into a downstream machine learning model for fraud detection.

The aforementioned two-stage methods treat account feature learning and fraud detection as separate processes, potentially hindering the framework's ability to effectively learn features directly relevant to the detection task. Additionally, the selection of classifiers significantly impacts the detection performance. To address these limitations, researchers have endeavored to develop end-to-end transaction graph learning frameworks for fraud detection.
Yu \etal~\cite{yu2021ponzi} construct transaction graphs and utilize graph convolutional networks (GCNs) to simultaneously learn transaction features and interaction behavior features of accounts, achieving end-to-end Ponzi scheme detection. 
Li \etal~\cite{li2022ttagn} design a tri-channel framework to separately capture the features of accounts, transactions, and transaction structures for fraud detection. 
Considering the large scale of transaction graph, Liu \etal~\cite{liu2022fa} reduce the graph size by filtering out low-importance accounts and use a two-layer GCN to learn the final account representations.
Liang \etal~\cite{liang2024ponziguard} analyze the code logic functions of contract accounts to construct a Contract Runtime Behavior Graph (CRBG) for describing the behavior of contract accounts. They employ a graph attention network (GAT)\cite{veličković2018graph} for account representation learning and fraud detection.

The aforementioned static methods fail to consider the dynamic nature of transactions when constructing transaction graphs. 
To address this, Jin \etal~\cite{jin2022dual} design a transaction graph construction method based on temporal evolution augmentation, incorporating time information while expanding the transaction graph scale, and ultimately combine transaction and code information for early detection of Ponzi schemes. 
Wang \etal~\cite{wang2023detecting} construct multiple subgraphs incorporating temporal information extracted from transaction records. By identifying and analyzing specific temporal motifs within the transaction graph, they capture account representations enriched with comprehensive temporal details, thereby enhancing the performance of existing fraud detection methods.
Zhang \etal~\cite{zhang2024grabphisher} construct temporal transaction graphs and combine LSTM and GCN for efficient learning of both temporal and structural transaction features.

Furthermore, the homogeneous transaction graphs constructed by the aforementioned methods neglect the heterogeneity of accounts and transaction behaviors. 
To address this, Zhou \etal~\cite{zhou2022behavior} incorporate call records from contract accounts as features for externally owned accounts and propose a hierarchical attention encoder based on contrastive learning to effectively learn account representations, alleviating the issue of label scarcity. 
Du \etal~\cite{du2023breaking} utilize a similar transaction graph construction method and design a GNN-based link prediction method to de-anonymize accounts in mixing services. 
Jin \etal~\cite{jin2022heterogeneous,10490264} predefine static and temporal meta-paths to describe the complex interaction behaviors in Ethereum and design a meta-path-based account feature augmentation module, which effectively improves the performance of existing Ponzi scheme detection methods. 
Considering that defining meta-paths requires expert knowledge and is time-consuming, Liu \etal~\cite{liu2022blockchain} and Xu \etal~\cite{xu2023illegal} employ transformer models to autonomously acquire meta-path patterns, subsequently aggregating neighbor accounts based on these meta-paths to obtain higher-order information for enhancing fraud detection.

\section{Ethereum Interaction Graph Modeling} \label{sec:pre}
\begin{figure*}[t]
    \centering
    \includegraphics[width=\textwidth]{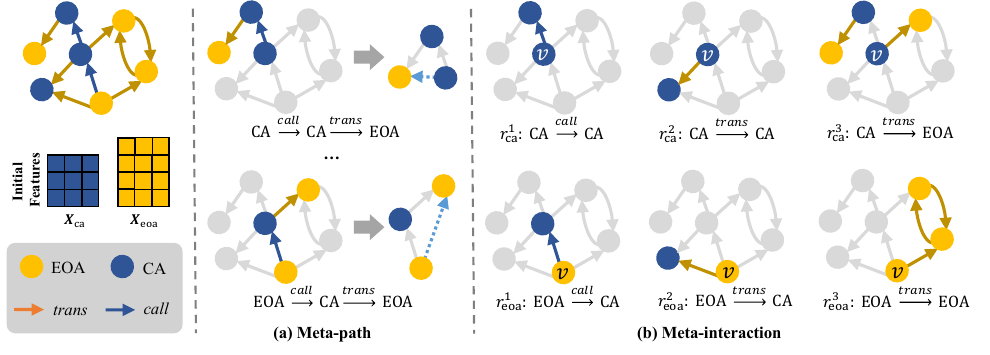}
    \caption{Illustration of meta-path and meta-interaction in heterogeneous Ethereum interaction graph.}
    \label{fig: meta}
\end{figure*}

\subsection{Preliminaries of Ethereum Data}\label{sec: Eth-klg}

In Ethereum, an account represents an entity that holds Ether and can be categorized into two types: Externally Owned Accounts (EOAs) and Contract Accounts (CAs). EOAs are managed by their respective private key holders, who possess the capability to initiate transactions on the Ethereum network. CAs are governed by their underlying smart contract code and can only be triggered to execute functions defined within the contract. Interactions between Ethereum accounts can be classified into two categories: transfer (\textit{trans}) and contract calls (\textit{call}). Transactions primarily involve the transfer of Ether, while contract calls obtain various services by triggering functions within the smart contract. Analyzing these interactions further can provide deeper insights into the functioning of the Ethereum network and uncover potential risks.

\subsection{Graph Modeling for Ethereum Fraud Detection}\label{sec: graph modeling}

In this paper, we analyze Ethereum interactions and perform fraud detection from a graph perspective. To better represent complex interaction scenarios, we model Ethereum data as a heterogeneous Ethereum interaction graph:
\begin{definition}[\textbf{Heterogeneous Ethereum Interaction Graph, HEIG}]
  Ethereum accounts and the interactions between them can be regarded as nodes and edges, respectively. By further refining their types, we construct a heterogeneous Ethereum interaction graph $G=(\mathcal{V}, \mathcal{E}, \mathcal{T}_\mathcal{V}, \mathcal{T}_\mathcal{E}, \boldsymbol{X}, \mathcal{Y})$, where $\mathcal{V}=\{v_1, v_2, \cdots, v_n\}$ is the set of account nodes, each with a specific type indicated by $\mathcal{T}_\mathcal{V}: v \mapsto \{\text{eoa}, \text{ca} \}$ as either EOA or CA, $\mathcal{E} = \{e_\textit{ij} \mid  e_\textit{ij} = (v_\textit{i}, v_\textit{j}), \ v_i, v_j \in \mathcal{V}\}$ is the set of interaction edges, each with a specific type indicated by $\mathcal{T}_\mathcal{E}: e \mapsto \{\text{trans}, \text{call} \}$ as either transfer or contract call. $\boldsymbol{X}=\left[\boldsymbol{x}_1, \boldsymbol{x}_2, \cdots, \boldsymbol{x}_\textit{n}\right]^\top \in \mathbb{R}^{n\times d}$ is the account feature matrix, and $d$ is the dimension of features. Due to the anonymity of Ethereum, the known account identity information is partially labeled, denoted as $\mathcal{Y}=\{ (v_\textit{i}, y_\textit{i})\mid v_\textit{i} \in \mathcal{V}_\textit{l},\  |\mathcal{V}_\textit{l}| \ll n \}$, where $\mathcal{V}_\textit{l}$ is the set of labeled accounts. 
\end{definition}

According to the above definition, each account $v_\textit{i}$ is assigned an initial feature vector $\boldsymbol{x}_\textit{i}$, which is often an indispensable input for most detection models. Considering that complex and well-designed features can be easily circumvented by fraudsters through behavior adjustment, thereby weakening the detection models and enabling evasion, we attempt to design a more general and concise set of manual features for account feature initialization. Specifically, we define the following manual features based on interaction details:
\begin{itemize}[leftmargin=10pt]
  \item Total / average investment and return under specific interaction types: $2 \times 2\times 2 = 8$ types.
  \item Balance under specific interaction types: $1\times 2=2$ types.
  \item Number of initiations and receptions under specific interaction types: $2\times 2=4$ types.
\end{itemize}
After statistics, we can construct a 14-dimensional feature vector for each account.
Furthermore, we model fraud detection in Ethereum as a node classification task on the graph. Specifically, we need to design a detector $f$ that characterizes the identity features of accounts by analyzing their interaction patterns, and ultimately establishes a mapping from accounts to identity labels: $f(v,G)\mapsto y$.

\begin{figure*}[!htb]
  \centering
  \includegraphics[width=\textwidth]{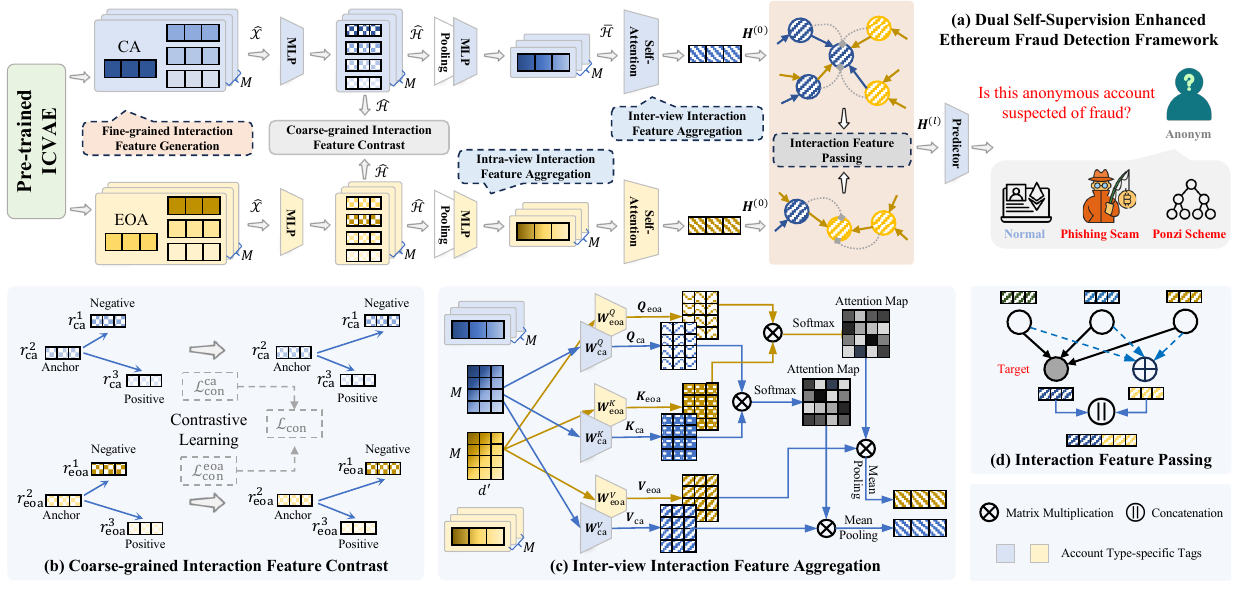}
      \caption{Illustration of the dual self-supervision enhanced Ethereum fraud detection framework (\model).}
      \label{fig: framework}
  \end{figure*}

\subsection{Refinement of Ethereum Interactions}

In Sec.~\ref{sec: Eth-klg}, we mention that interactions in Ethereum can be categorized into two types: transfer (\textit{trans}) and contract calls (\textit{call}), which we refer to as coarse-grained interactions. 
Several studies~\cite{jin2022heterogeneous,10490264} have defined meta-paths to describe the behavior patterns of Ethereum accounts, as shown in Fig.~\ref{fig: meta}(a). While such practice is feasible, it encounters several issues. Firstly, the definition of meta-paths relies on statistical analysis and expert knowledge, which limits their generalizability. Secondly, specific meta-paths define specific behavior patterns, and their adaptability may weaken when confronted with complex and variable fraud scenarios. Finally, as the scale of interactions increases, the number of meta-path instances will surge, potentially losing behavioral representativeness.

To describe Ethereum interactions in a more fine-grained manner, we further refine the interactions based on the mechanism of Ethereum, where transfers can occur between any accounts while contract calls can only target contract accounts. Specifically, in conjunction with the structure of \graph, we propose the concept of meta-interaction:
\begin{definition}[\textbf{Meta-interactions}]
  In an interaction graph, meta-interaction represents the most fine-grained interaction behavior pattern between arbitrary entity pairs $\langle s,t \rangle$, which can be formalized as $r = \langle\mathcal{T}_\mathcal{V}(s), \mathcal{T}_\mathcal{E}(e), \mathcal{T}_\mathcal{V}(t)\rangle = \mathcal{T}_\mathcal{V}(s) \stackrel{\mathcal{T}_\mathcal{E}(e)}{\longrightarrow }\mathcal{T}_\mathcal{V}(t)$, where $s$ and $t$ denote the initiator and recipient of the interaction respectively, and $e$ denotes the specific interaction between $s$ and $t$.
\end{definition}

Unlike the coarse-grained interactions \textit{trans} and \textit{call} which only reflect the ways in which accounts interact with each other, meta-interactions further describe the ways in which specific types of accounts can interact with each other, and thus can be referred to as fine-grained interactions. Based on the definition of meta-interaction and the characteristics of Ethereum, we can identify six types of meta-interactions from \graph, as shown in Fig.~\ref{fig: meta}(b).
Furthermore, we define $\mathcal{R}=\{r^1_\text{ca}, r^2_\text{ca}, r^3_\text{ca}, r^1_\text{eoa}, r^2_\text{eoa}, r^3_\text{eoa}\}$ to denote the set of meta-interactions, and the detailed description is shown in Table.~\ref{tb:Meta-interactions}.

\begin{table} 
  \centering
  \caption{Descriptions of different meta-interactions.}
    \resizebox{\linewidth}{!}{
      \renewcommand\arraystretch{1.5}
  \begin{tabular}{ll} 
  \hline\hline
  Meta-interactions         & Descriptions               \\ 
  \hline
    $r^1_\text{ca}:  \text{CA} \stackrel{\text{\textit{call}}}{\longrightarrow } \text{CA}$  & \makecell[l]{Interaction between CA to achieve complex logical functions \\or extend functionalities.}  \\
    $r^1_\text{eoa}: \text{EOA} \stackrel{\text{\textit{call}}}{\longrightarrow} \text{CA}$  & \makecell[l]{EOA initiates a request to call contract functions or update\\ contract states.}          \\
    $r^2_\text{ca}:  \text{CA}  \stackrel{\text{\textit{trans}}}{\longrightarrow}\text{CA}$  & \makecell[l]{CA handles automated asset management based on predefined \\internal logic.}              \\
    $r^2_\text{eoa}: \text{EOA}\stackrel{\text{\textit{trans}}}{\longrightarrow} \text{CA}$  & \makecell[l]{EOA transfers Ether or tokens to CA to provide funding or \\trigger its execution logic.}  \\
    $r^3_\text{ca}:  \text{CA} \stackrel{\text{\textit{trans}}}{\longrightarrow} \text{EOA}$ & \makecell[l]{CA transfers Ether or tokens to an EOA, distributing funds \\or rewards.}                 \\
    $r^3_\text{eoa}: \text{EOA} \stackrel{\text{\textit{trans}}}{\longrightarrow}\text{EOA}$ & \makecell[l]{Direct Ether or token transfers between EOAs without \\involving smart contract logic.}   \\
  \hline\hline
  \end{tabular}}
  \label{tb:Meta-interactions}
\end{table}

\section{Methodology} \label{sec: method}
In this section, we discuss the design details of the proposed framework \model, as schematically depicted in Fig.~\ref{fig: framework}. This framework takes \graph ~as input and outputs the prediction labels for the target accounts, and it primarily consists of the following modules:
1) A fine-grained interaction feature generation module to simulate and augment the behavioral features of accounts;
2) A multi-view feature learning module to deeply characterize the behavior patterns of accounts;
3) A coarse-grained interaction feature contrast module to enhance the framework's ability to distinguish different account behaviors.
Next, we introduce the details of each module.

\subsection{Fine-grained Ethereum Interaction Feature Generation}

\begin{figure}[t]
  \centering
  \includegraphics[width=0.9\linewidth]{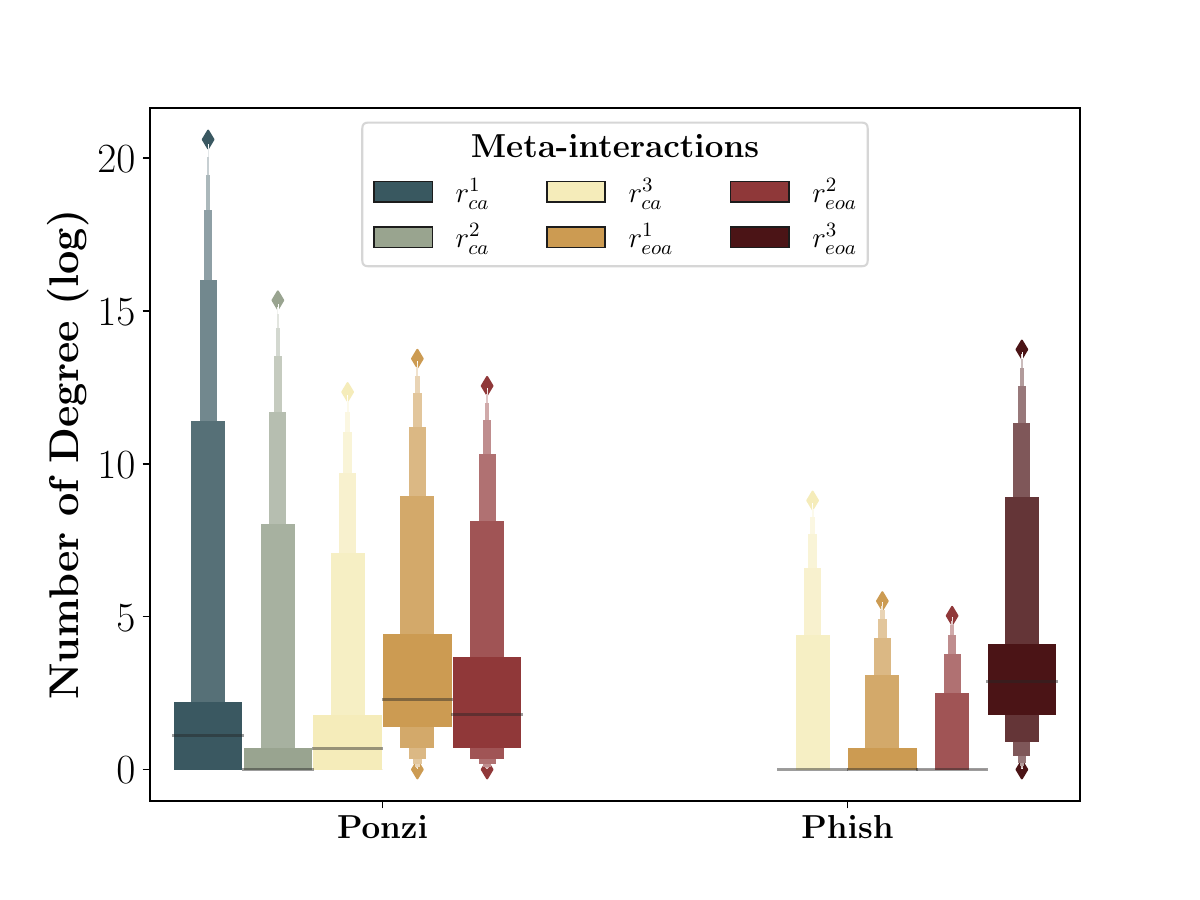}
  \caption{Illustration of the imbalanced distribution in account interaction types. Boxes indicate the degree correlation statistics for labeled accounts across different interaction types.}
  \label{fig: box}
\end{figure}

In complex Ethereum interaction scenarios, account entities may encounter the following issues: 1) different accounts exhibit varying levels of interaction activity, leading to imbalanced degree distributions, as illustrated in Fig.~\ref{fig: degree-dis}, which reveals typical long-tail distributions in both datasets. Specifically, the number of low-degree (inactive) accounts significantly exceeds that of high-degree (active) accounts, thus forming the classical degree imbalanced problem in complex networks; 
2) the same account participates in different types of interactions to varying extents, resulting in imbalanced interaction distributions, as illustrated in Fig.~\ref{fig: box}, where box plot represents the statistical distribution of account degrees across various interactions, including the minimum, upper quartile, median, lower quartile and the maximum value.
Long and thin boxes indicate a large difference in the degree distribution of accounts under this interaction type. The variations in box shapes between different interaction types suggest greater differences in degree distributions between interaction types.
These diverse and imbalanced interaction phenomena can introduce biases when characterizing account interaction behaviors.
To alleviate these issues, we introduce a generative self-supervised mechanism to augment the interaction features of Ethereum accounts, particularly for those with low activity levels and low-frequency interaction types.

\subsubsection{\textbf{Model Architecture and Pre-training}}
Specifically, we propose a fine-grained Ethereum interaction feature generation module, as illustrated in Fig.~\ref{fig: ICVAE}.
Considering that for a target account $v_\textit{i}$, the characteristics of its neighboring entities $\boldsymbol{x}_\textit{j}$ in its local interaction environment are influenced by its own characteristics $\boldsymbol{x}_\textit{i}$ as well as its interaction habits $e_\textit{ij}$. Based on this prior, if we want to augment more interaction features for the target account, we have to understand the potential feature distribution of its local interaction environment. Assuming that the potential features of the local interaction environment of account $v_\textit{i}$ follow a prior distribution $\boldsymbol{z} \sim p_\theta\left(\boldsymbol{z} \mid \boldsymbol{x}_\textit{i}, \boldsymbol{r}_\textit{ij}\right)$, where $\boldsymbol{r}_\textit{ij}$ is the one-hot encoded form of the meta-interaction between $v_\textit{i}$ and $v_\textit{j}$, as shown in Fig.~\ref{fig: ICVAE}(a), then the neighbor features $\boldsymbol{x}_\textit{j}$ follow a generative distribution $\boldsymbol{x}_\textit{j} \sim p_\theta\left(\boldsymbol{x} \mid \boldsymbol{z}, \boldsymbol{x}_\textit{i}, \boldsymbol{r}_\textit{ij}\right)$. Further, we design an interaction-aware conditional variational autoencoder (\textit{ICVAE}) to learn the interaction feature distribution of target accounts and achieve fine-grained interaction feature generation. 
Compared to the conditional variational autoencoder (CVAE)~\cite{liu2022local} with a single condition setting, our \textit{ICVAE} utilizes both the target account and specific meta-interaction as conditions, and neighbor as input to train the encoder and decoder.
It serves as a generative model with a log-likelihood probability of $\log p_\theta\left(\boldsymbol{x}_\textit{j} \mid \boldsymbol{x}_\textit{i}, \boldsymbol{r}_\textit{ij}\right)$, which denotes the probability that the model generates the observed neighborhood feature conditional on the target account feature $\boldsymbol{x}_\textit{i}$ and the meta-interaction feature $\boldsymbol{r}_\textit{ij}$. Let $\theta$ and $\phi$ represent the generative parameters and variational parameters respectively, we have:
\begin{equation}
  \small
  \begin{aligned}
    & \log p_\theta\left(\boldsymbol{x}_\textit{j} \mid \boldsymbol{x}_\textit{i}, \boldsymbol{r}_\textit{ij}\right) \\
    & =\mathbb{E}_{q_\phi\left(\boldsymbol{z} \mid \boldsymbol{x}_\textit{j}, \boldsymbol{x}_\textit{i}, \boldsymbol{r}_\textit{ij} \right)}\left[\log p_\theta \left(\boldsymbol{x}_\textit{j}\mid \boldsymbol{x}_\textit{i}, \boldsymbol{r}_\textit{ij}\right)\right]\\
    & =\mathbb{E}_{q_\phi\left(\boldsymbol{z} \mid \boldsymbol{x}_\textit{j}, \boldsymbol{x}_\textit{i}, \boldsymbol{r}_\textit{ij} \right)}\left[\log \frac{p_\theta\left(\boldsymbol{x}_\textit{j}, \boldsymbol{z} \mid \boldsymbol{x}_\textit{i}, \boldsymbol{r}_\textit{ij}\right)}{p_\theta\left(\boldsymbol{z} \mid \boldsymbol{x}_\textit{j}, \boldsymbol{x}_\textit{i}, \boldsymbol{r}_\textit{ij}\right)}\right]\\
    & =\mathbb{E}_{q_\phi\left(\boldsymbol{z} \mid \boldsymbol{x}_\textit{j}, \boldsymbol{x}_\textit{i}, \boldsymbol{r}_\textit{ij} \right)}\left[\log \frac{p_\theta\left(\boldsymbol{x}_\textit{j}, \boldsymbol{z} \mid \boldsymbol{x}_\textit{i}, \boldsymbol{r}_\textit{ij}\right)}{q_\phi\left(\boldsymbol{z} \mid \boldsymbol{x}_\textit{j}, \boldsymbol{x}_\textit{i}, \boldsymbol{r}_\textit{ij}\right)}\cdot\frac{q_\phi\left(\boldsymbol{z} \mid \boldsymbol{x}_\textit{j}, \boldsymbol{x}_\textit{i}, \boldsymbol{r}_\textit{ij}\right)}{p_\theta\left(\boldsymbol{z} \mid \boldsymbol{x}_\textit{j}, \boldsymbol{x}_\textit{i}, \boldsymbol{r}_\textit{ij}\right)}\right] \\
    & =\mathbb{E}_{q_\phi\left(\boldsymbol{z} \mid \boldsymbol{x}_\textit{j}, \boldsymbol{x}_\textit{i}, \boldsymbol{r}_\textit{ij} \right)}\left[\log p_\theta\left(\boldsymbol{x}_\textit{j}, \boldsymbol{z} \mid \boldsymbol{x}_\textit{i}, \boldsymbol{r}_\textit{ij}\right) - q_\phi\left(\boldsymbol{z} \mid \boldsymbol{x}_\textit{j}, \boldsymbol{x}_\textit{i}, \boldsymbol{r}_\textit{ij}\right) \right]    \\
    & \qquad\qquad\qquad + \mathbb{E}_{q_\phi\left(\boldsymbol{z} \mid \boldsymbol{x}_\textit{j}, \boldsymbol{x}_\textit{i}, \boldsymbol{r}_\textit{ij} \right)}\left[\log \frac{q_\phi\left(\boldsymbol{z} \mid \boldsymbol{x}_\textit{j}, \boldsymbol{x}_\textit{i}, \boldsymbol{r}_\textit{ij}\right)}{p_\theta\left(\boldsymbol{z} \mid \boldsymbol{x}_\textit{j}, \boldsymbol{x}_\textit{i}, \boldsymbol{r}_\textit{ij}\right)} \right]\\
    & =\mathbb{E}_{q_\phi\left(\boldsymbol{z} \mid \boldsymbol{x}_\textit{j}, \boldsymbol{x}_\textit{i}, \boldsymbol{r}_\textit{ij} \right)}\left[\log p_\theta\left(\boldsymbol{x}_\textit{j}, \boldsymbol{z} \mid \boldsymbol{x}_\textit{i}, \boldsymbol{r}_\textit{ij}\right) - q_\phi\left(\boldsymbol{z} \mid \boldsymbol{x}_\textit{j}, \boldsymbol{x}_\textit{i}, \boldsymbol{r}_\textit{ij}\right) \right]    \\
    & \qquad\qquad\qquad + \text{KL}\left(q_\phi\left(\boldsymbol{z} \mid \boldsymbol{x}_\textit{j}, \boldsymbol{x}_\textit{i}, \boldsymbol{r}_\textit{ij}\right) \parallel p_\theta\left(\boldsymbol{z} \mid \boldsymbol{x}_\textit{j}, \boldsymbol{x}_\textit{i}, \boldsymbol{r}_\textit{ij}\right) \right)\\
    & \geq\mathbb{E}_{q_\phi\left(\boldsymbol{z} \mid \boldsymbol{x}_\textit{j}, \boldsymbol{x}_\textit{i}, \boldsymbol{r}_\textit{ij} \right)}\left[\log p_\theta\left(\boldsymbol{x}_\textit{j}, \boldsymbol{z} \mid \boldsymbol{x}_\textit{i}, \boldsymbol{r}_\textit{ij}\right) - q_\phi\left(\boldsymbol{z} \mid \boldsymbol{x}_\textit{j}, \boldsymbol{x}_\textit{i}, \boldsymbol{r}_\textit{ij}\right) \right]    \\
    & = \mathbb{E}_{q_\phi\left(\boldsymbol{z} \mid \boldsymbol{x}_\textit{j}, \boldsymbol{x}_\textit{i}, \boldsymbol{r}_\textit{ij} \right)}[\log p_\theta\left( \boldsymbol{z} \mid \boldsymbol{x}_\textit{i}, \boldsymbol{r}_\textit{ij}\right) + \log p_\theta\left(\boldsymbol{x}_\textit{j} \mid \boldsymbol{z}, \boldsymbol{x}_\textit{i}, \boldsymbol{r}_\textit{ij}\right)  \\
    & \qquad\qquad\qquad - q_\phi\left(\boldsymbol{z} \mid \boldsymbol{x}_\textit{j}, \boldsymbol{x}_\textit{i}, \boldsymbol{r}_\textit{ij}\right)] \\
    & =\mathbb{E}_{q_\phi\left(\boldsymbol{z} \mid \boldsymbol{x}_\textit{j}, \boldsymbol{x}_\textit{i}, \boldsymbol{r}_\textit{ij} \right)}\left[\log p_\theta\left(\boldsymbol{x}_\textit{j} \mid \boldsymbol{z}, \boldsymbol{x}_\textit{i}, \boldsymbol{r}_\textit{ij}\right)\right] \\
    & \qquad\qquad\qquad - \mathbb{E}_{q_\phi\left(\boldsymbol{z} \mid \boldsymbol{x}_\textit{j}, \boldsymbol{x}_\textit{i}, \boldsymbol{r}_\textit{ij} \right)}\left[\log \frac{q_\phi\left(\boldsymbol{z} \mid \boldsymbol{x}_\textit{j}, \boldsymbol{x}_\textit{i}, \boldsymbol{r}_\textit{ij}\right)}{p_\theta\left( \boldsymbol{z} \mid \boldsymbol{x}_\textit{i}, \boldsymbol{r}_\textit{ij}\right)}\right] \\
    & =\mathbb{E}_{q_\phi\left(\boldsymbol{z} \mid \boldsymbol{x}_\textit{j}, \boldsymbol{x}_\textit{i}, \boldsymbol{r}_\textit{ij} \right)}\left[\log p_\theta\left(\boldsymbol{x}_\textit{j} \mid \boldsymbol{z}, \boldsymbol{x}_\textit{i}, \boldsymbol{r}_\textit{ij}\right)\right] \\
    & \qquad\qquad\qquad  -\text{KL}\left(q_\phi\left(\boldsymbol{z} \mid \boldsymbol{x}_\textit{j}, \boldsymbol{x}_\textit{i}, \boldsymbol{r}_\textit{ij}\right) \parallel p_\theta\left( \boldsymbol{z} \mid \boldsymbol{x}_\textit{i}, \boldsymbol{r}_\textit{ij}\right) \right)\\
    & = \text{ELBO} 
    \end{aligned}
\end{equation}
where $q_\phi\left(\boldsymbol{z} \mid \boldsymbol{x}_\textit{j}, \boldsymbol{x}_\textit{i}, \boldsymbol{r}_\textit{ij}\right)$ denotes the variational posterior distribution and is used to approximate the true posterior distribution $p_\theta\left(\boldsymbol{z} \mid \boldsymbol{x}_\textit{j}, \boldsymbol{x}_\textit{i}, \boldsymbol{r}_\textit{ij}\right)$, $\text{KL}(q \parallel p)$ denotes the Kullback-Leibler divergence between the two distributions $q$ and $p$.
Since directly computing the log-likelihood probability is difficult, we approximate it by maximizing its lower bound (\ie Evidence Lower Bound, ELBO). Thus, the ELBO-based optimization objective of our \textit{ICVAE} can be derived as follows:
\begin{figure}[!htb]
  \centering
  \includegraphics[width=\linewidth]{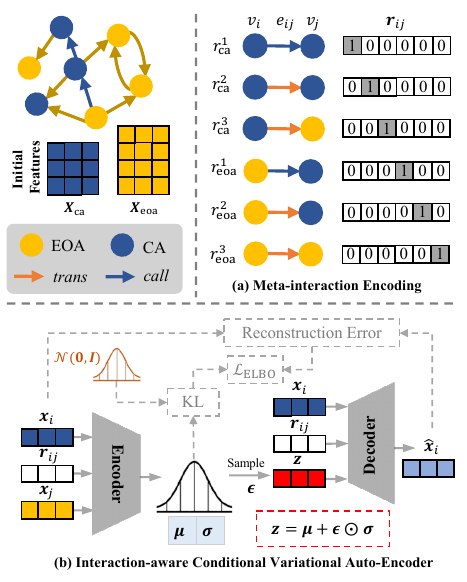}
      \caption{Illustration of the fine-grained Ethereum interaction feature generation module.}
    \label{fig: ICVAE}
\end{figure}
\begin{equation}
  \begin{aligned}
    & \mathcal{L}_\text{ELBO} = \frac{1}{|\mathcal{N}_{v_\textit{i}}|} \sum_{v\in \mathcal{N}_{v_\textit{i}}} \log p_\theta\left(\boldsymbol{x}_\textit{j} \mid \boldsymbol{z}^{(v)}, \boldsymbol{x}_\textit{i}, \boldsymbol{r}_\textit{ij}\right) \\
    & \qquad\qquad  -\text{KL}\left(q_\phi\left(\boldsymbol{z} \mid \boldsymbol{x}_\textit{j}, \boldsymbol{x}_\textit{i}, \boldsymbol{r}_\textit{ij}\right) \parallel p_\theta\left( \boldsymbol{z} \mid \boldsymbol{x}_\textit{i}, \boldsymbol{r}_\textit{ij}\right) \right)
  \end{aligned}
\end{equation}
where $\mathcal{N}_{v_\textit{i}}$ represents the neighbor set of account $v_\textit{i}$. 
The first term in $\mathcal{L}_\text{BLEO}$ is the reconstruction error, which measures the difference between the features generated by \textit{ICVAE} and the original interaction features. The second term is the KL divergence, which measures the difference between the variational posterior distribution and the prior distribution.

After specifying the optimization objective, we are ready to pre-train an \textit{ICVAE} for all accounts. The \textit{ICVAE} consists of an encoder and a decoder, as illustrated in Fig.~\ref{fig: ICVAE}(b). During the training phase, it takes interaction feature pairs $\left\{\left(\boldsymbol{x}_\textit{i}, \boldsymbol{x}_\textit{j}, \boldsymbol{r}_\textit{ij}\right) \mid v_\textit{i}\in \mathcal{V}, v_\textit{j}\in\mathcal{N}_{v_\textit{i}}\right\}$ as input to maximize the $\mathcal{L}_\text{ELBO}$. Specifically, the encoder maps the conditions $\left(\boldsymbol{x}_\textit{i}, \boldsymbol{r}_\textit{ij}\right)$ and the input data $\boldsymbol{x}_\textit{j}$ to the latent space, learning and outputting the distribution parameters of the latent variables $\boldsymbol{z}$:
\begin{equation}
  \boldsymbol{\mu}, \boldsymbol{\sigma} = \text{Encoder}\left(\boldsymbol{x}_\textit{i}, \boldsymbol{x}_\textit{j},\boldsymbol{r}_\textit{ij}\right)
\end{equation}
where the mean $\boldsymbol{\mu}$ and standard deviation $\boldsymbol{\sigma}$ are used to describe the variational posterior distribution $q_\phi\left(\boldsymbol{z} \mid \boldsymbol{x}_\textit{j}, \boldsymbol{x}_\textit{i}, \boldsymbol{r}_\textit{ij}\right)$.
The decoder takes the conditions $\left(\boldsymbol{x}_\textit{i}, \boldsymbol{r}_\textit{ij}\right)$ and the sampled latent variable $\boldsymbol{z}$ as inputs, generating the interaction features $\hat{\boldsymbol{x}_\textit{i}}$ associated with account $v_\textit{i}$:
\begin{equation}\label{eq: decoder}
  \begin{array}{c}
    \boldsymbol{z} = \boldsymbol{\mu} + \boldsymbol{\epsilon}  \odot  \boldsymbol{\sigma}\\
    \hat{\boldsymbol{x}}_\textit{i} = \text{Decoder}\left( \boldsymbol{x}_\textit{i}, \boldsymbol{r}_\textit{ij}, \boldsymbol{z}\right) 
  \end{array}
\end{equation}
where $\boldsymbol{z}$ is sampled via re-parameterization trick, $\boldsymbol{\epsilon} \sim \mathcal{N}(\boldsymbol{0}, \boldsymbol{I})$ is Gaussian noise that follows a standard normal distribution, and $\odot$ is the element-wise multiplication.
By maximizing $\mathcal{L}_\text{BLEO}$, \textit{ICVAE} can learn the optimal parameters for both the encoder and the decoder, thereby minimizing the reconstruction error of the interaction features under given conditions, and ensuring that the posterior distribution parameterized by the encoder closely approximates the prior distribution.

\begin{algorithm}[t]
\caption{The training process of \model.}
\begin{algorithmic}[1]
\label{al:1}
\REQUIRE Heterogeneous Ethereum interaction graph $G=(\mathcal{V}, \mathcal{E}, \mathcal{T}_\mathcal{V}, \mathcal{T}_\mathcal{E},\boldsymbol{X},\mathcal{Y})$, meta-interaction set $\mathcal{R}$, the number of views $M$, and pre-trained ICVAE. 
\ENSURE \model. 
\STATE Load pre-trained ICVAE model;
\FOR{each account $v_\textit{i} \in \mathcal{V}$}
    \STATE Get feature vector $\boldsymbol{x}_\textit{i}$;
    \WHILE{$M > 0$}
    \FOR{each neighbor account $v_\textit{j} \in \mathcal{N}_{v_\textit{i}}$}
        \FOR{different meta-interaction $r_\textit{ij} \in \mathcal{R}$}
        \STATE Generate interaction features $\hat{\boldsymbol{x}}_\textit{i}$ via Eq.~(\ref{eq: decoder});
        \ENDFOR
    \ENDFOR
    \STATE $M \gets M-1$;
    \ENDWHILE
\ENDFOR
\STATE Obtain augmented interaction feature $\hat{\mathcal{X}}$;
\STATE Perform type-specific transformation via Eq.~(\ref{eq:type-specific mapping});
\STATE Calculate coarse-grained interaction feature contrast loss $\mathcal{L}_\text{con}^\ast$ via Eq.~(\ref{eq: contrast})
\STATE Perform intra-view feature aggregation via Eq.~(\ref{eq:intra-view aggr});
\STATE Perform inter-view feature aggregation via Eq.~(\ref{eq:meanpooling});
\STATE Obtaining final account representation $\boldsymbol{H}$ with two-layer message passing according to Eq.~(\ref{eq:message passing});
\STATE Calculate classification loss via Eq.~(\ref{eq: classification});
\STATE Train \model via minimize the optimization objective in Eq.~(\ref{eq: final loss});
\RETURN \model.
\end{algorithmic}
\end{algorithm}

\subsubsection{\textbf{Feature Generation}}
After the pre-training phase, we utilize \textit{ICVAE} to generate additional interaction features for the target accounts. These generated interaction features can further enrich the neighborhood information of the target accounts and alleviate the issue of feature distribution imbalance. Specifically, given a target account $v_\textit{i}$ and different meta-interactions $r\in\mathcal{R}$ as conditions, \textit{ICVAE} can generate rich interaction features for it:
\begin{equation}\label{eq: aug_feat}
  \hat{\mathcal{X}}_\textit{i}=
    \left\{\boldsymbol{x}_\textit{i}, \hat{\boldsymbol{x}}_\textit{i}^{r^1_\ast}, \hat{\boldsymbol{x}}_\textit{i}^{r^2_\ast}, \hat{\boldsymbol{x}}_\textit{i}^{r^3_\ast}\right\}^{M}
\end{equation}
where $\ast=\mathcal{T}_\mathcal{V}(v_\textit{i})$ indicates the type of $v_\textit{i}$, and $\hat{\boldsymbol{x}}_\textit{i}^{\textit{r}^1_\ast}$ is the feature generated conditional on $(\boldsymbol{x}_\textit{i}, \boldsymbol{r}^1_\ast)$. 
Note that under specific conditions, we can repeatedly sample the latent variable $\boldsymbol{z}$ to generate interaction features multiple times, \ie repeatedly utilizing \textit{ICVAE} to generate multiple interaction features under the same conditions, forming multiple feature views, and further alleviating the feature distribution imbalance issue. The number of multiple feature views is denoted by $M$.

\subsection{Multi-view Ethereum Interaction Feature Learning}

With the pre-trained \textit{ICVAE}, we obtain augmented interaction features from multiple views, which can alleviate the issue of sparse interaction features for several accounts to some extent. To better characterize the account behavior patterns, the multi-view interaction feature learning module will process both the initial and generated interaction features to uncover potential fraud risks. This module will encode and fuse the multi-view interaction features and ultimately propagate the features by combining the structural information of \graph.
The design details of the module are described as follows, and the complete process is presented in Algorithm~\ref{al:1}.

\subsubsection{\textbf{Input Feature Encoding}}\label{sec: input encode}
\begin{figure}[t] 
  \centering
  \includegraphics[width=\linewidth]{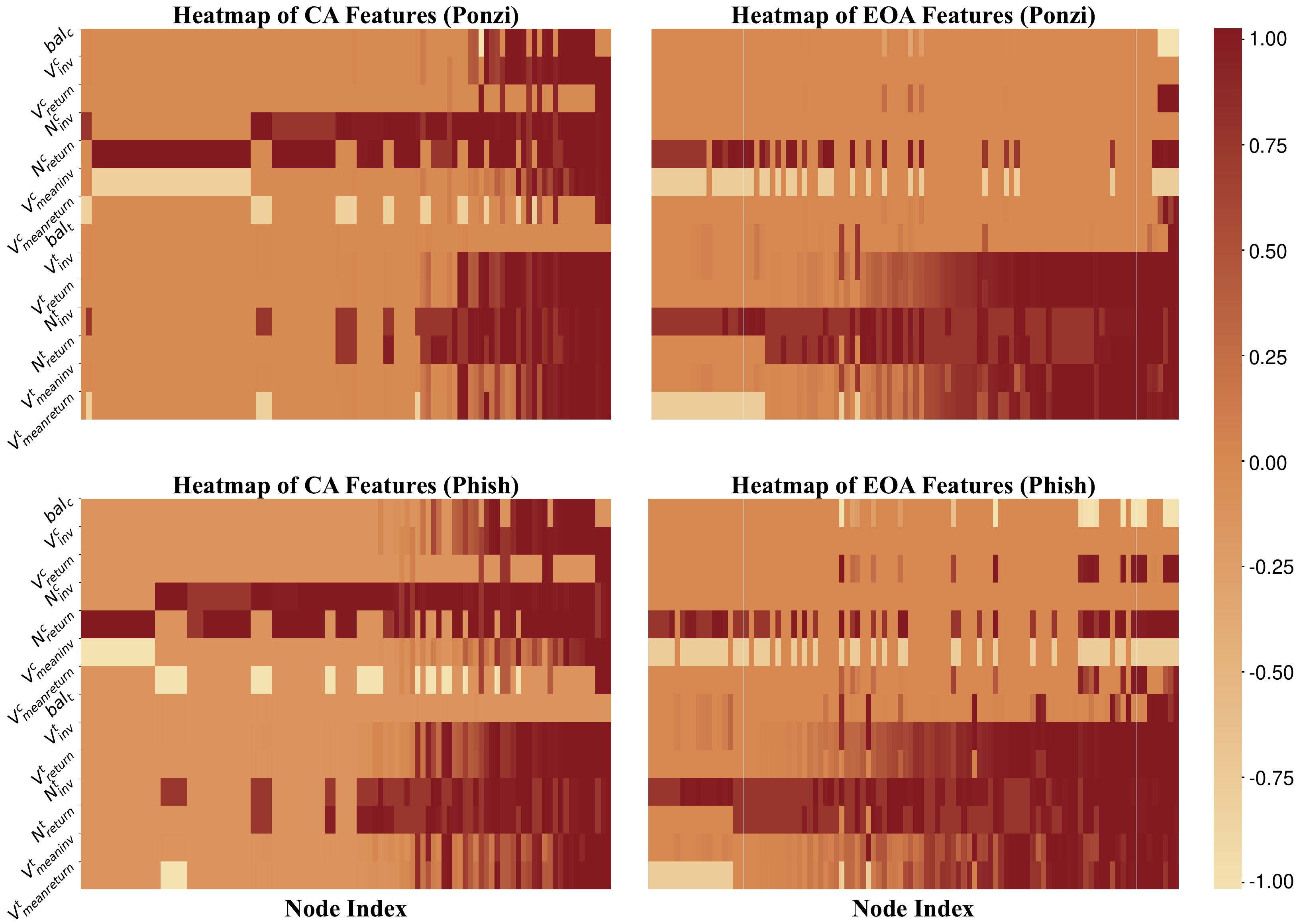}
    \caption{Illustrate the feature distribution of different account variations in various fraud datasets.}
    \label{fig: heatmap}
\end{figure}

The multi-view interaction features generated by the \textit{ICVAE} remain in the input space, with their dimensions represented as $\hat{\mathcal{X}}_\textit{i}\in\mathbb{R}^{M \times 4 \times d}$. To further explore the underlying information in the interaction features, a common practice is to design a weight-sharing encoder for all target accounts, which accepts the interaction features of the accounts as input and outputs high-dimensional hidden features. However, we consider such practice to be inappropriate for the following reasons. 
We sample a certain number of EOA and CA and count their 14-dimensional normalized features, as shown in Fig.~\ref{fig: heatmap}, where the first 7-dimensional features are derived from the \textit{call}, and the last 7-dimensional features are derived from the \textit{trans}. 
We can observe noticeable differences among the features of accounts within the same type, as well as among different types of accounts.
For example, there is a significant difference in the \textit{call} features extracted from EOA and CA, and the \textit{trans} features of EOA are relatively larger than those of CA.
In such cases, a weight-sharing transformation usually fails to adequately distinguish the differences among accounts.
Therefore, to more adaptively characterize the interaction preferences of different types of accounts, we design account type-specific input feature encoding components, which follows a weight separation setting~\cite{NCGNN,gong2023neighborhood}.
Specifically, for the multi-view interaction features of target account $v_\textit{i}$ in the input space, we map them into the hidden space using an MLP parameterized by $\hat{\mathbf{\Theta}}_\ast \in \mathbb{R}^{d\times d'}$, obtaining the hidden representation $\hat{\mathcal{H}}_\textit{i}\in\mathbb{R}^{M\times 4\times d'}$ as follows:
\begin{equation} \label{eq:type-specific mapping}
  \hat{\mathcal{H}}_\textit{i}=\text{tanh}\left(\hat{\mathcal{X}}_\textit{i}\cdot\hat{\mathbf{\Theta}}_\ast\right) = \left\{\boldsymbol{h}_\textit{i}, \hat{\boldsymbol{h}}_\textit{i}^{r^1_\ast}, \hat{\boldsymbol{h}}_\textit{i}^{r^2_\ast}, \hat{\boldsymbol{h}}_\textit{i}^{r^3_\ast}\right\}^{M}
\end{equation}
where $\ast = \mathcal{T}_\mathcal{V}(v_\textit{i})$ indicates that the MLP performs an account type-specific transformation on the input features.

\subsubsection{\textbf{Intra-view and Inter-view Feature Aggregation}}
Subsequently, since \textit{ICVAE} can generate three sets of augmented interaction features for each account, the feature dimensions within each view will be tripled. To reduce the parameter scale of our framework and improve training efficiency while preventing overfitting, we further perform \textbf{intra-view interaction feature aggregation} via mean pooling, followed by an MLP to ensure the expressive power of the aggregated features:
\begin{equation} \label{eq:intra-view aggr}
  \begin{aligned}
    \bar{\mathcal{H}}_\textit{i}&=\text{tanh}\left(\operatorname{MeanPooling}\left(\hat{\mathcal{H}}_\textit{i}\right)\cdot\bar{\mathbf{\Theta}}_\ast\right) \\
    &=\text{tanh}\left(\frac{1}{4}\left(\boldsymbol{h}_\textit{i}+ \hat{\boldsymbol{h}}_\textit{i}^{r^1_\ast}+ \hat{\boldsymbol{h}}_\textit{i}^{r^2_\ast}+ \hat{\boldsymbol{h}}_\textit{i}^{r^3_\ast}\right)\cdot\bar{\mathbf{\Theta}}_\ast\right)
  \end{aligned}
\end{equation}
where $\bar{\mathcal{H}}_\textit{i}\in\mathbb{R}^{M\times d'}$ and the account type-specific weight $\bar{\mathbf{\Theta}}_\ast \in \mathbb{R}^{d' \times d'}$ is utilized to parameterize the MLP.

After intra-view feature aggregation, the multi-view interaction features will be dimensionally reduced and updated, which is beneficial for the training of \model. Furthermore, we consider that the interaction features from different views exhibit certain correlations and complementarities, contributing to the final account feature representation to varying degrees. Therefore, we introduce the self-attention mechanism to capture the interdependencies among different views and achieve \textbf{inter-view interaction feature aggregation}, as shown in Fig.~\ref{fig: framework}(c).
Specifically, in order to use the self-attention mechanism to compute correlations between different views, we initially generate query, key, and value vectors and subsequently integrate these features by calculating the attention weights. We define three type-specific transformation weights $\boldsymbol{W}^\textit{Q}_\ast$, $\boldsymbol{W}^\textit{K}_\ast$, $\boldsymbol{W}^\textit{V}_\ast \in \mathbb{R}^{d' \times d'}$ to generate the three vectors:
\begin{equation} \label{eq:self-attention}
  \boldsymbol{Q}_\ast = \bar{\mathcal{H}}_\textit{i} \cdot \boldsymbol{W}^\textit{Q}_\ast, \quad
  \boldsymbol{K}_\ast = \bar{\mathcal{H}}_\textit{i} \cdot \boldsymbol{W}^\textit{K}_\ast, \quad
  \boldsymbol{V}_\ast = \bar{\mathcal{H}}_\textit{i} \cdot \boldsymbol{W}^\textit{V}_\ast
\end{equation}
where $\boldsymbol{Q}_\ast$ is used to find relevant features, $\boldsymbol{K}_\ast$ represents the critical information of the features, and $\boldsymbol{V}_\ast$ is the feature representation used for weighted summation. 
Next, we calculate the inter-view correlations, also known as attention, to obtain the comprehensive feature representation of each view, and finally fuse the features from all views:
\begin{equation} \label{eq:meanpooling}
  \bar{\boldsymbol{H}}_\textit{i} = \operatorname{MeanPooling}\left(\operatorname{Softmax}\left(\boldsymbol{Q}\boldsymbol{K}^\top\right)\cdot\boldsymbol{V}\right)
\end{equation}
Note that during the inter-view feature aggregation process, the query vector of each view $\boldsymbol{Q}_\textit{a}$ performs dot-product with the key vectors of other views $\boldsymbol{K}_\textit{b}$ to obtain relevance score, \ie $\boldsymbol{Q}_\textit{a}\boldsymbol{K}^\top_\textit{b}$, which indicates the importance of the features in view $a$ to those in view $b$. Using the attention scores across all views after softmax normalization, we perform a weighted summation of the value vectors to obtain the comprehensive feature representation for each view: $\sum_{b=1}^{M} \operatorname{Softmax}\left(\boldsymbol{Q}_\textit{a}\boldsymbol{K}^\top_\textit{b}\right)\cdot \boldsymbol{V}_\textit{b}$. Finally, we fuse the features of all views via mean pooling to obtain the feature representation of the target account $\bar{\boldsymbol{H}}_\textit{i}$.

\subsubsection{\textbf{Interaction Feature Passing}}
The behavioral patterns of accounts are not only related to their own characteristics, but also influenced by the characteristics of the objects they interact with. In order to more accurately characterize the behavioral patterns of the target accounts, we incorporate the interaction structure information among accounts in \graph ~and obtain the final account behavior representation through interaction feature passing. Specifically, we initially aggregate the features of the interaction objects associated with the target account, subsequently propagate them to the target account, and ultimately perform a deep transformation on the aggregated features. The above feature passing process will be iterated $l$ times to complete the final account feature learning.
\begin{equation} \label{eq:message passing}
  \boldsymbol{H}_\textit{i}^{(l)} = \text{tanh} \left(\left(\ \sum_{j\in \mathcal{N}_{v_\textit{i}}} \boldsymbol{H}_\textit{j}^{(l-1)} \right) \parallel \boldsymbol{H}_\textit{i}^{(l-1)}   \right) \cdot \mathbf{\Theta}_\ast^{(l-1)}
\end{equation}
where $\boldsymbol{H}_\textit{i}^{(0)}=\bar{\boldsymbol{H}}_\textit{i}$ and $\mathbf{\Theta}_\ast^{(l-1)}$ denotes the transformation weight in the feature passing process at the $l-1$-th iteration.

\begin{table*}[t]
  \centering
  \caption{Statistics of Ethereum fraud detection datasets. $|\mathcal{V}|$ represents the number of accounts, $\text{Num.}~ r$ represents the number of meta-interaction instances, label ratio represents the quantity ratio between fraud samples and normal samples.}
  \label{tb:dataset}
  \resizebox{\textwidth}{!}{
    \renewcommand\arraystretch{1.5}
  \begin{tabular}{c|ccccccccc} 
  \hline\hline
  {Datasets}      & $|\mathcal{V}_\text{ca}|$  & $|\mathcal{V}_\text{eoa}|$  & $\text{Num.}~ r_\text{ca}^1$ & $\text{Num.}~ r_\text{ca}^2$ & $\text{Num.}~ r_\text{ca}^3$ & $\text{Num.}~ r_\text{eoa}^1$ & $\text{Num.}~ r_\text{eoa}^2$ & $\text{Num.}~ r_\text{eoa}^3$ & Label Ratio \\ 
  \hline
  Ponzi  & 3,104,188 & 25,199,827 & 4,900,340  & 618,572   & 2,423,011  & 25,606,952 & 3,337,998  & 32,559,586 & 191:1151\\ 
  Phish   & 2,279,349 & 22,429,276 & 1,879,385  & 527,653   & 2,096,037  & 20,425,382 & 2,994,136  & 26,293,882 & 1206:1557\\
  \hline\hline
  \end{tabular}}
\end{table*}
\subsection{Coarse-grained Ethereum Interaction Feature Contrast} \label{sec: contrast}
In the definition of meta-interaction, we combine the type of interactions and the type of targets to distinguish fine-grained interaction behaviors. 
And each type of account involves up to three different meta-interactions. 
Take EOA as an example, this type of account involves three meta-interactions: $r_\text{eoa}^1$, $r_\text{eoa}^2$ and $r_\text{eoa}^3$, where $r_\text{eoa}^1$ involves contract call, and $r_\text{eoa}^2$ and $r_\text{eoa}^3$ involve transferring money. When the EOA makes a transfer, we consider that the CA or EOA that receives the transfer plays a similar role; whereas when the EOA makes a contract call and a transfer, even if the target is the same CA, the CA plays different roles in these two types of interactions. Therefore, the interaction type should play a dominant role in characterizing the interaction behavior. For this purpose, we use interaction type as one of the conditions for generating interaction features when designing \textit{ICVAE}. Considering the limitations of MLP in encoding input features discussed in Sec.~\ref{sec: input encode}, we further introduce a contrastive self-supervised mechanism to enhance its ability to distinguish different interaction features, proposing the \textbf{coarse-grained Ethereum interaction feature contrast}, as shown in Fig.~\ref{fig: framework}(b).

Specifically, from a coarse-grained interaction perspective, we believe that $r_\ast^2$ and $r_\ast^3$ should have more similar representations because they both involve transfer behaviors, while $r_\ast^2$ and $r_\ast^1$ should exhibit differences since the latter involves contract call behavior. Therefore, for a target account $v_\textit{i}$ of type $\ast=\mathcal{T}_\mathcal{V}(v_\textit{i})$, we consider meta-interaction $r_\ast^2$ as anchor sample, $r_\ast^3$ as positive sample, and $r_\ast^1$ as negative sample.
During framework training, we expect to make the representations of anchor and positive samples more similar and those of anchor and negative samples more distinct, which can be achieved by minimizing the following objective:
\begin{equation}\label{eq: contrast}
    \mathcal{L}_\text{con}^\ast = \sum_{v_\textit{i} \in \mathcal{V}_\ast}\ \max\left\{0,\| \hat{\boldsymbol{h}}_\textit{i}^{r_\ast^2}-\hat{\boldsymbol{h}}_\textit{i}^{r_\ast^3} \|_2^2 -\| \hat{\boldsymbol{h}}_\textit{i}^{r_\ast^2} - \hat{\boldsymbol{h}}_\textit{i}^{r_\ast^1} \|_2^2 +\alpha\right\} 
\end{equation}
where $\alpha$ is a margin hyperparameter that controls the minimum difference between positive and negative samples.
The above optimization objective can serve as regularization, facilitating the framework in capturing the similarities between interactions of the same type and the differences between interactions of different types, thereby enhancing the framework's ability to distinguish various interaction behaviors.

\subsection{Framework Training} \label{sec: training}
In this paper, the \textit{ICVAE} model used for generating interaction features can be pre-trained by maximizing the $\mathcal{L}_\text{ELBO}$. And Ethereum fraud detection is modeled as a node classification task on \graph. By applying a prediction head $f_\Psi $ to the final account representations, we map them into the corresponding prediction labels and compute the classification loss based on cross-entropy:
\begin{equation}\label{eq: classification}
  \mathcal{L}_\text{pred}=-\frac{1}{N} \sum_{i=1}^N y_i \cdot \log \left(f_\Psi\left(\boldsymbol{H}_\textit{i}\right)\right)
\end{equation}
where $N$ is the number of accounts in a batch (\ie batch size).
Furthermore, the coarse-grained interaction feature contrast module can be regarded as an auxiliary task, serving as a regularization for the account classification task to jointly optimize the entire framework. The optimization objective of joint training can be defined as:
\begin{equation}\label{eq: final loss}
  \mathcal{L} = \mathcal{L}_\text{pred} + \lambda \cdot (\mathcal{L}_\text{con}^\text{eoa} + \mathcal{L}_\text{con}^\text{ca})
\end{equation}
where $\lambda$ serves as a trade-off coefficient to control the contribution of the feature contrast module.

\section{Experiments} \label{sec:exp}
\subsection{Data Collection}

We collect a certain amount of labeled data from the \textit{Xblock}\footnote[3]{\url{http://xblock.pro}} and \textit{Etherscan}\footnote[4]{\url{https://cn.etherscan.com}} platforms, covering 191 Ponzi accounts and 1151 non-Ponzi accounts, as well as 1206 phishing accounts and 1557 non-phishing accounts. Based on these labeled accounts, we further extract their first- and second-order transaction objects to construct the \graph, obtaining two fraud detection datasets with imbalanced label distribution. 
The detailed statistics of these datasets is presence in Table~\ref{tb:dataset}.
According to the statistics, we observe that although the Ponzi dataset contains fewer labeled instances, it exhibits significantly richer interaction behaviors. Specifically, the total number of interactions in the Ponzi dataset exceeds that of the Phish dataset by 15,229,984, with an additional 8,192,525 \textit{call} interactions. This discrepancy arises because executing Ponzi fraud requires multiple contract triggers, whereas phishing fraud in the Phish dataset is typically completed by merely guiding users to perform a single transfer.

\subsection{Comparison Methods}
In this paper, we choose four categories of methods for comparison: machine learning-based method (LightGBM~\cite{xu2024ewdps}), homogeneous graph analysis-based methods (Trans2Vec~\cite{wu2020phishers}, GCN~\cite{yu2021ponzi}, GAT~\cite{liang2024ponziguard}, SAGE~\cite{tharani2024unified}, FA-GNN~\cite{liu2022fa}), heterogeneous graph analysis-based methods (HGT~\cite{hu2020heterogeneous}, RGCN~\cite{schlichtkrull2018modeling}, ieHGCN~\cite{yang2021interpretable}, HTSGCN~\cite{huang2023ethereum}), and sequence analysis-based methods (BERT4ETH~\cite{hu2023bert4eth}, ZipZap~\cite{hu2024zipzap}). 
We exclude meta-path-based methods from our comparison due to the significant effort required in designing meta-paths for targeting different fraud behaviors.

\begin{table*}[t]
  \centering
  \caption{The results of fraud detection in terms of Precision(\%), Recall(\%), Binary-F1(\%), Micro-F1(\%), Macro-F1(\%) and Standard Deviation(\%). The best results are highlighted in bold, while the second-best results are underlined. ``Rank'' indicates the performance ranking of a method on a single fraud dataset, ``Average Rank'' indicates the average performance ranking across all datasets, and ``Group Rank'' indicates the average performance ranking of a category of methods across all datasets.}
  \label{tb: all result}
  \resizebox{\linewidth}{!}{
  \renewcommand\arraystretch{1.5}
\begin{tabular}{c|cccccc|cccccc|c|c} 
\hline\hline
\multirow{2}{*}{Methods} & \multicolumn{6}{c|}{Ponzi Scheme Dataset}                                                                                                      & \multicolumn{6}{c|}{Phish Scam Dataset}                                                                                                        & \multirow{2}{*}{\begin{tabular}[c]{@{}c@{}}Average\\Rank\end{tabular}} & \multirow{2}{*}{\begin{tabular}[c]{@{}c@{}}Group\\Rank\end{tabular}}  \\
                         & Precision               & Recall                  & Binary-F1               & Micro-F1                & Macro-F1                & Rank         & Precision               & Recall                  & Binary-F1               & Micro-F1                & Macro-F1                & Rank         &                                                                        &                                                                       \\ 
\hline
LightGBM~                & 72.29$\pm$11.91         & 38.42$\pm$2.11          & 49.97$\pm$4.18          & 89.07 $\pm$1.30         & 71.92$\pm$2.46          & 8.2          & 94.13$\pm$ 0.67         & \uline{97.01$\pm$1.13}  & 95.54$\pm$0.37          & 96.06 $\pm$0.31         & 96.00$\pm$0.32          & 4.8          & 6.5                                                                    & 6.50                                                                  \\ 
\hdashline
Trans2Vec~               & \textbf{91.16$\pm$3.24} & 48.42$\pm$4.59 & 63.12$\pm$4.08          & \uline{92.04$\pm$0.69}  & 79.33$\pm$2.22          & 2.8          & 77.21$\pm$1.85          & 74.19$\pm$2.11          & 75.63$\pm$0.96         & 79.17$\pm$0.83       & 78.71$\pm$0.82          & 13.0         & 7.9                                                                    & \multirow{5}{*}{8.14}                                                 \\
GCN~                     & 67.61$\pm$8.70          & 27.89$\pm$6.78          & 38.44$\pm$5.22          & 87.66$\pm$0.49          & 65.78$\pm$2.51          & 12.2         & 92.05$\pm$1.35          & 88.38$\pm$1.14          & 90.18$\pm$1.19          & 91.61$\pm$1.03          & 91.43$\pm$1.05          & 11.6         & 11.9                                                                   &                                                                       \\
GAT~                     & 77.62$\pm$6.48          & 47.37$\pm$6.66          & 58.40$\pm$5.01          & 90.56$\pm$0.80          & 76.53$\pm$2.68          & 5.0          & 90.20$\pm$1.52          & 95.19$\pm$0.86          & 92.62$\pm$0.95          & 93.38$\pm$0.90          & 93.31$\pm$0.90          & 9.6          & 7.3                                                                    &                                                                       \\
SAGE~                    & 70.88$\pm$7.06          & 37.37$\pm$3.07          & 48.78$\pm$3.26          & 88.92$\pm$0.79          & 71.29$\pm$1.82          & 9.6         & 93.46$\pm$0.45          & 96.02$\pm$0.62          & 94.72$\pm$0.44          & 95.33$\pm$0.39          & 95.27$\pm$0.39          & 6.2          & 7.9                                                                    &                                                                       \\
FA-GNN~                  & 66.69$\pm$6.00          & 42.11$\pm$4.40          & 51.17$\pm$1.76          & 88.70$\pm$0.50          & 72.39$\pm$0.76          & 9.0          & 96.92$\pm$0.20          & 96.60$\pm$0.31          & \uline{96.76$\pm$0.21}  & \uline{97.18$\pm$0.18}  & \uline{97.13$\pm$0.19}  & 2.6          & 5.8                                                                    &                                                                       \\ 
\hdashline
HGT~                     & 58.47$\pm$5.69          & 45.26$\pm$1.97          & 50.83$\pm$2.26          & 87.58$\pm$1.26          & 71.86$\pm$1.49          & 10.4         & 90.89$\pm$2.16          & 92.20$\pm$1.16          & 91.53$\pm$1.33          & 92.55$\pm$1.22          & 92.44$\pm$1.22          & 11.0         & 10.7                                                                   & \multirow{4}{*}{7.95}                                                 \\
RGCN~                    & 69.61$\pm$5.64          & 53.16$\pm$5.10          & 60.26$\pm$5.29          & 90.11$\pm$1.26          & 77.31$\pm$3.00          & 5.6          & 94.95$\pm$1.57          & 94.36$\pm$1.07          & 94.64$\pm$0.45          & 95.33$\pm$0.43          & 95.25$\pm$0.43          & 6.4          & 6.0                                                                    &                                                                       \\
ieHGCN~                  & 61.49$\pm$5.34          & 38.42$\pm$11.36         & 46.59$\pm$10.69         & 88.03$\pm$1.21          & 69.92$\pm$5.62          & 11.4         & 94.38$\pm$0.83          & 94.36$\pm$3.63          & 94.32$\pm$1.75          & 95.08$\pm$1.39          & 94.99$\pm$1.45          & 7.4          & 9.4                                                                    &                                                                       \\
HTSGCN~                  & 67.47$\pm$5.79          & 42.11$\pm$3.33          & 51.59$\pm$2.34          & 88.85$\pm$0.78          & 72.64$\pm$1.30          & 8.2          & 97.00$\pm$0.68          & 96.27$\pm$0.95          & 96.63$\pm$0.37          & 97.07$\pm$0.31          & 97.02$\pm$0.32          & 3.2          & 5.8                                                                    &                                                                       \\ 
\hdashline
BERT4ETH                 & 72.83$\pm$3.75          & \textbf{65.26$\pm$4.21} & \uline{68.67$\pm$2.18}  & 91.60$\pm$0.60          & \uline{81.91$\pm$1.23}  & 2.4          & 92.84$\pm$1.64          & 92.95$\pm$1.17          & 92.87$\pm$0.42          & 93.78$\pm$0.42          & 93.68$\pm$0.41          & 9.2          & 5.8                                                                    & \multirow{2}{*}{5.15}                                                 \\
ZipZap                   & 69.64$\pm$8.44          & 56.22$\pm$4.65          & 61.77$\pm$3.68          & 90.37$\pm$1.28          & 78.13$\pm$2.17          & 4.6          & \textbf{98.27$\pm$1.24} & 94.22$\pm$1.61          & 96.19$\pm$0.47          & 96.96$\pm$0.35          & 96.83$\pm$0.37          & 4.4          & 4.5                                                                    &                                                                       \\ 
\hdashline
Meta-IFD                 & \uline{79.71$\pm$3.66}  & \uline{61.58$\pm$1.29}  & \textbf{69.46$\pm$2.05} & \textbf{92.34$\pm$0.60} & \textbf{82.54$\pm$1.20} & \textbf{1.4} & \uline{97.37$\pm$0.55}  & \textbf{98.01$\pm$0.17} & \textbf{97.68$\pm$0.27} & \textbf{97.97$\pm$0.24} & \textbf{97.94$\pm$0.24} & \textbf{1.2} & \textbf{1.3}                                                           & \textbf{1.30}                                                         \\
\hline\hline
\end{tabular}}
\end{table*}
\subsection{Experimental Settings}

In our framework, for the pre-trained generation module, the input is \graph, the encoder is a 3-layer MLP with output dimensions \{32, 64, 50\}, and the decoder is a 2-layer MLP with output dimensions \{32, 14\}, the learning rate is set to 0.01. For the contrastive module, the margin hyperparameter $\alpha$ is set to 0.3. For the multi-view feature learning module, the search space for the number of views, hidden dimensions, and learning rate is \{2, 3, 4\}, \{32, 64, 128\}, and \{0.01, 0.001\} respectively, the number of attention heads is set to 4. During training, the search space for the trade-off coefficient $\lambda$ is \{0.001, 0.01, 0.1, 1\}. The graph structure input for feature passing is the sampled \graph.

For heterogeneous graph analysis-based methods, the inputed heterogeneous graph is obtained by sampling two layers of neighbors of the target nodes, with each layer sampling 100 neighbors based on different interaction behaviors. For homogeneous graph analysis-based methods, the sampled homogeneous graph is obtained by removing type information from the heterogeneous graph. Both methods employ a two-layer structure, with the search space for hidden dimensions and learning rate being \{32, 64, 128\} and \{0.01, 0.001\}, respectively. For methods involving attention, the number of attention heads is set to 4. For the two-stage Trans2Vec, we set the window size, walk length, and the number of walks per account to 10, 5, and 20, respectively, with the downstream classifier being an SVM with default parameters.
For LightGBM, the search spaces for the n\_estimators parameter and learning rate are \{100, 200, 300\} and \{0.01, 0.001\}.

For sequence analysis-based methods, during the pre-trained phase, we utilize default parameters with Transformer layers set to 8, hidden dimension set to 64, number of attention heads set to 2, maximum sequence length set to 100, and the optimal learning rate selected from \{0.001, 0.0001\}. During the classification phase, a two-layer MLP is employed with an optimal combination of learning rate chosen from \{0.01, 0.001\} and hidden dimensions selected from \{32, 64, 128\}, respectively.
The parameter configuration ensures that the search ranges for hyperparameters align with those employed in the comparative detection methods discussed in this paper, thereby maintaining the fairness of the experimental comparison.

For all datasets, we divide them into training, validation, testing sets in a proportion of 6:2:2.
To prevent overfitting, we set an early stop with a patience of 50.
All experiments will be run 5 times, with the average performance and standard deviation reported. 
Performance metrics used are Precision, Recall, Binary-F1, Micro-F1 and Macro-F1.
The first three metrics evaluate the detection performance for positive samples with the binary evaluation. 
Micro-F1 emphasizes the overall performance across all categories, akin to Accuracy in binary classification. 
Macro-F1 calculates the Binary-F1 metrics of both positive and negative samples separately and reports the average results.
Combining the above metrics, we can evaluate the model's detection ability for each sample type.

\subsection{Evaluation on Fraud Detection}
To evaluate the effectiveness and superiority of our \model, we compare it against 12 baselines on two fraud datasets. The detection results are reported in Table~\ref{tb: all result}, from which we can draw the following conclusions.

Overall, compared with the optimal machine learning, homogeneous graph analysis, heterogeneous graph analysis methods, and sequence analysis-based methods, our \model ~achieves 14.77\%, 4.05\%, 6.76\%, and 0.77\% improvement in detecting Ponzi schemes, and 2.02\%, 0.83\%, 0.95\%, and 1.15\% in detecting phishing scams (Macro-F1 metric), which demonstrates the effectiveness and superiority of our \model in Ethereum fraud detection.

To evaluate the performance in detecting fraudulent accounts, Table~\ref{tb: all result} presents the precision, recall, and F1-score under binary evaluation metrics. As observed, our \model achieves the best performance in terms of F1-score and ranks second in both precision and recall. Taking Ponzi detection as an example, Trans2Vec performs well in precision but poorly in recall, indicating that this method is overly conservative in detecting fraud, leading to a high rate of missed detections. Conversely, BERT4ETH performs well in recall but poorly in precision, suggesting that this approach tends to misclassify normal accounts as fraudulent, resulting in a high false positive rate. The F1-score balances these two aspects, and the best F1 performance demonstrates that our \model provides a more stable performance in fraud detection.

The detection performance of different types of baselines varies significantly across various fraud categories. 
We select representative baselines for analysis.
LightGBM achieves relatively powerful performance in phishing detection but performs poorly in Ponzi scheme detection, indicating that the general manual features defined by machine learning methods cannot adequately describe diverse fraudulent behaviors. To achieve better detection performance, machine learning methods necessitate meticulous design of manual features tailored to different fraud categories, which lacks good generalization and heavily depends on expert knowledge. The two-stage homogeneous graph analysis methods represented by Trans2Vec suffer from similar limitations, in addition to the fact that it focuses more on the structural information of the transaction graph.
Other major research in this field face similar challenges. For example, BERT4ETH, a more recent advancement that incorporates Transformer models for fraud detection, demonstrates high performance on Ponzi detection but exhibits poor generalizability when detecting diverse fraudulent behaviors. The ZipZap method, an improvement based on BERT4ETH, achieves some enhancement in generalizability but at the cost of reduced performance in certain aspects. This indicates that achieving both high generalizability and high effectiveness in Ethereum fraud detection remains a significant technical challenge.
Other end-to-end heterogeneous graph analysis methods show moderate detection performance for different fraudulent behaviors, somewhat surpassing homogeneous graph analysis methods. 
A reasonable explanation is that both are affected by the interaction graph construction and data sampling methods. Finally, our \model achieves the best detection performance for different frauds, indicating its generalizability and effectiveness in fraud detection.

\subsection{Analysis of Multi-view Feature Learning Module}
We first deploy ablation and parameter analysis to comprehensively investigate the effectiveness of the multi-view interaction feature learning module.
\subsubsection{Impact of View Setting}
\begin{figure}[!htb] 
  \centering
  \includegraphics[width=\linewidth]{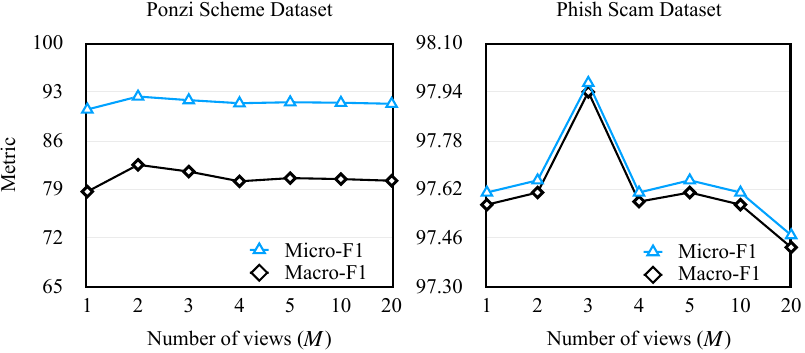}
    \caption{Impact on the number of views ($M$).}
    \label{fig: views}
\end{figure}

As shown in Table~\ref{tb: ablation}, when our \model ~is adapted to a single-view setting, \ie ``Module Removal ($w/o \ M$)'', its performance in detecting Ponzi schemes and phishing scams decreases by 4.89\% and 0.34\% (Macro-F1 metric), respectively, indicating that the utilization of multi-view settings can enhance the characterization of account features to some extent. Furthermore, under the multi-view setting, we vary the number of views within \{1, 2, 3, 4, 5, 10, 20\}
, as depicted in Fig.~\ref{fig: views}, from which we can observe that:
The framework's performance (measured by Macro-F1 metric) shows a general trend of initially increasing and then decreasing with the number of views. This suggests that increasing the number of views appropriately is beneficial for the detection of minority class fraud samples, but too many views may lead to feature redundancy, which fails to improve the detection performance and even in turn interferes with analysis of complex Ethereum interaction behaviors.

\begin{table}[t]
  \centering
  \caption{Performance comparison between Meta-IFD and its ablation models.}
  \resizebox{\linewidth}{!}{
    \renewcommand\arraystretch{1.5}
    \begin{tabular}{ll|cc|cc} 
\hline\hline
\multicolumn{2}{c|}{\multirow{2}{*}{Methods}}                                                                 & \multicolumn{2}{c|}{Ponzi Scheme Dataset}                 & \multicolumn{2}{c}{Phish Scam Dataset}                      \\
\multicolumn{2}{l|}{}                                                                                         & Micro-F1                    & Macro-F1                    & Micro-F1                     & Macro-F1                     \\ 
\hline
\multirow{3}{*}{\begin{tabular}[c]{@{}l@{}}Module\\Removal\end{tabular}}    & $w/o \ I$                         & 91.67$\pm$0.87 & 80.13$\pm$1.82 & 97.61$\pm$0.45  & 97.58$\pm$0.45  \\
& $w/o\ M$                         & 90.48$\pm$1.07 & 78.69$\pm$1.29 & 97.61$\pm$0.39  & 97.57$\pm$0.39  \\
& $w/o\ \mathcal{L}_\text{con}$    & 91.23$\pm$0.73 & 80.42$\pm$1.33 & 97.76$\pm$0.56  & 97.72$\pm$0.56  \\ 
\hdashline
\multirow{3}{*}{\begin{tabular}[c]{@{}l@{}}Multi-view\\Aggregation\end{tabular}} & \textit{Concat}    & 91.15$\pm$0.28 & 79.59$\pm$0.89 & 97.61$\pm$0.31 & 97.58$\pm$0.31  \\        & \textit{Sum}     & 91.23$\pm$0.38 & 79.91$\pm$1.06 & 97.65$\pm$0.30  & 97.61$\pm$0.31  \\
           & \textit{Naive-atten} & 91.67$\pm$1.01 & 81.25$\pm$2.10 & 97.72$\pm$0.52  & 97.69$\pm$0.53  \\ 
\hdashline
\multirow{2}{*}{Features}             & \textit{call}                 & 88.18$\pm$0.86 & 73.18$\pm$0.97 & 84.30$\pm$0.14  & 83.76$\pm$0.12  \\
 & \textit{trans}                & 89.96$\pm$0.53 & 77.33$\pm$0.91 & 96.60$\pm$0.21  & 96.55$\pm$0.22  \\ 
\hdashline
Others           & \textit{Sharing}       & 92.19$\pm$0.53 & 81.40$\pm$1.49 & 97.32$\pm$0.39  & 97.28$\pm$0.39  \\
\hdashline
\multicolumn{2}{c|}{Meta-IFD}                 & \textbf{92.34$\pm$0.60} & \textbf{82.54$\pm$1.20} & \textbf{97.94$\pm$0.29}  & \textbf{97.90$\pm$0.30}  \\
\hline\hline
\end{tabular}}
      \label{tb: ablation}
\end{table}

\subsubsection{Impact of Weight Separation Setting}
To validate the effectiveness of the account type-specific transformation in our \model, we compare the performance of the framework under a weight-sharing setting, \ie ``Others (\textit{Sharing})'',
as shown in Table~\ref{tb: ablation}. The results show that weight separation learning for different types of account features yields positive gains, suggesting an enhanced distinguishability among the features of distinct accounts. 

\subsubsection{Impact of Self-attention Setting}
To validate the superiority of the self-attention in inter-view feature aggregation, we compare three aggregation operations: summation, concatenation, and naive attention. As shown in Table~\ref{tb: ablation}, the self-attention aggregation operation facilitates our \model ~in achieving improvement ranging from 1.59\% to 3.71\% and 0.22\% to 0.33\% on the two datasets (Macro-F1 metric), respectively. This result demonstrates that the self-attention mechanism can more effectively extract and fuse key features from different views, thereby enhancing fraud detection.

\begin{figure*}
  \centering
  \includegraphics[width=\textwidth]{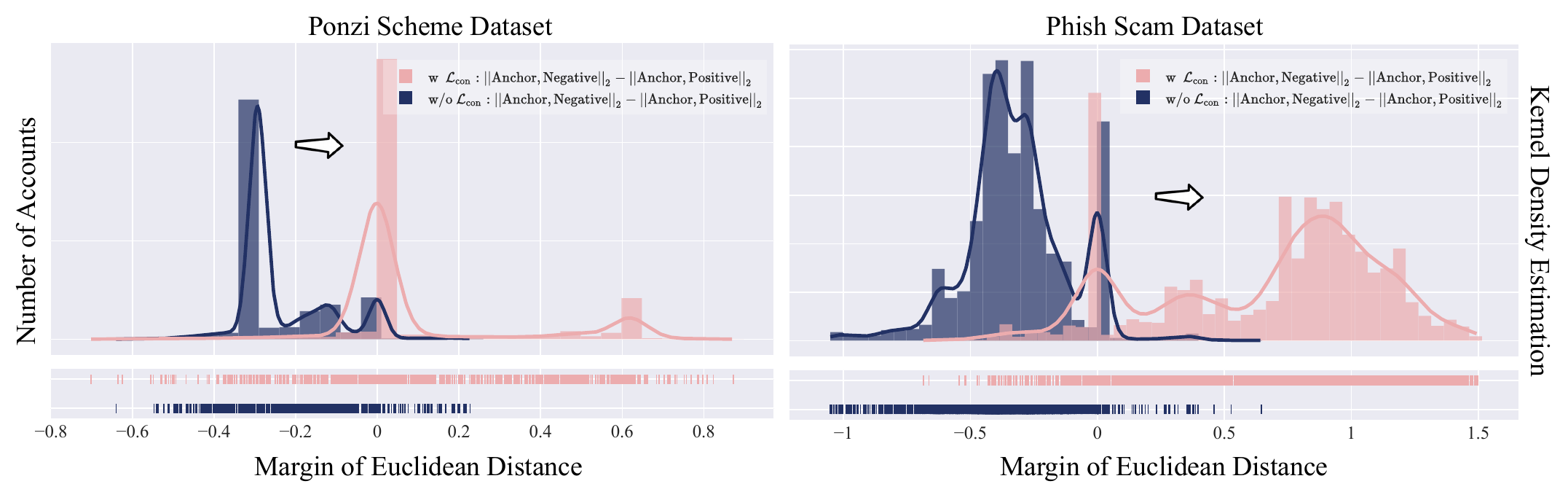}
    \caption{Illustration of the meta-interaction feature distance margin distribution before and after introducing contrastive self-supervision.
    The bars denote the distribution of the number of accounts, with their width signifying the size of the statistical window. Curves represent trends in the distribution.}
    \label{fig: contract}
\end{figure*}

\begin{figure}
  \centering
  \includegraphics[width=\linewidth]{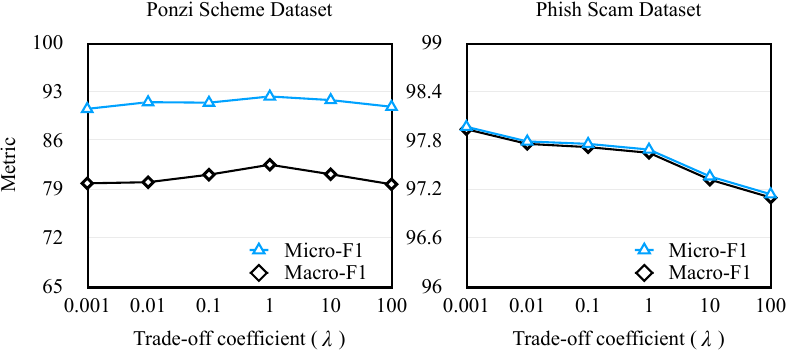}
    \caption{Impact of the trade-off coefficient ($\lambda$).}
    \label{fig:lambda}
\end{figure}

\subsection{Analysis of Coarse-grained Feature Contrast Module}
Furthermore, we conduct ablation and parameter experiments to analyze the effectiveness of the coarse-grained interaction feature contrast module.

\subsubsection{Effect of Contrastive Self-supervision}
As shown in Table~\ref{tb: ablation}, the removal of the contrast module results in a performance degradation of 2.64\% and 0.18\% on the two datasets, respectively. This observation highlights the effective improvement of our framework's performance in fraud detection through feature contrast. Additionally, we compare the distance margin distribution between different types of meta-interaction features before and after introducing contrastive self-supervision, as shown in Fig.~\ref{fig: contract}. 
We utilize the Euclidean distance to quantify the similarity between samples, assessing the discrepancy in similarity scores between positive and negative samples. Subsequently, we record both the count of accounts falling within the specified similarity window and their corresponding trend.
It can be observed that after introducing contrastive self-supervision, the peak of the distance margin distribution shifts significantly to the right, indicating that the distance between anchors $r^2_\ast$ and positive samples $r^3_\ast$ is reduced, while the distance between anchors and negative samples $r^1_\ast$ is increased. This phenomenon further illustrates that introducing contrastive self-supervision facilitates \model ~in learning the similarities among interactions of the same type, as well as distinctions between interactions of different types, improving its ability to distinguish different interaction behaviors and ultimately enhancing fraud detection.

\subsubsection{Impact of Trade-off Coefficient}
Furthermore, we investigate how the framework's performance is affected by the intensity of the auxiliary module, which is controlled by the trade-off coefficient $\lambda$. We vary $\lambda$ within \{0.001, 0.01, 0.1, 1, 10, 100\} and observe the fluctuations in the framework's performance, as shown in Fig.~\ref{fig:lambda}. Firstly, we can see that in detecting Ponzi schemes, \model ~achieves better detection results when $\lambda$ is larger. Combined with the data statistics from Table~\ref{tb:dataset} and the distance margin distribution in Fig.~\ref{fig: contract}, a reasonable explanation is that the Ponzi scheme dataset is more severely affected by data distribution imbalance and label scarcity issues, and higher intensity of contrast self-supervised constraint are needed to help \model ~better distinguish different interaction behavior patterns. Therefore, appropriately increasing the $\lambda$ setting helps improve the framework's detection performance.
For the phishing scam dataset, its data distribution is relatively balanced, and the label size is relatively abundant compared to the Ponzi scheme dataset, which facilitates effective learning of account behavior representations within \model. Consequently, only moderate contrastive self-supervision constraint is required to enhance the discriminative capability of our framework.

\subsection{Analysis of Feature Generation}

\subsubsection{Impact of Input Features}
To validate the effectiveness and superiority of the handcrafted features proposed in this work, we separately extract features based on \textit{call} and \textit{trans} interactions, \ie we retain only the 7-dimensional features derived from a single interaction type for subsequent fraud detection. For fairness, we extend these features to the same 14 dimensions. The experimental results are shown in Table~\ref{tb: ablation} under ``Feature (\textit{call})'' and ``Feature (\textit{trans})''. It can be observed that the Macro-F1 score drops by 6.74\% and 1.40\%, respectively.
Notably, detection using ``Feature (\textit{trans})'' outperforms that using ``Feature (\textit{call})'', particularly in phishing detection. This is because phishing fraud typically occurs between EOAs, involving fewer \textit{call} behaviors. Furthermore, \textit{trans}, as a more common behavior, carries richer semantic information.
In summary, considering different interaction behaviors during feature construction enables the acquisition of a more comprehensive feature set, thereby improving the effectiveness of subsequent fraud detection.

\subsubsection{Impact of Using Interaction Types as Conditions in Feature Generation}
To validate the effectiveness of incorporating interaction types as conditions during the feature generation process, we remove the interaction-aware component, degrading ICVAE into a standard CVAE model for interaction feature generation. The fraud detection performance of this ablation model is presented in Table.~\ref{tb: ablation} under ``Module Removal $(w/o \  I)$''. The results show that the macro-F1 scores drop by 3.01\% and 0.33\%, respectively.
Notably, detecting Ponzi schemes is more dependent on interaction information, as Ponzi schemes often exploit more interaction behaviors to achieve their fraudulent objectives. Therefore, incorporating interaction information during the pretraining phase effectively captures fine-grained behavioral patterns of accounts.
In conclusion, integrating interaction types as conditions during the feature generation stage allows for the creation of richer features for low-degree nodes, thereby improving subsequent detection performance.

\begin{figure}[t] 
  \centering
  \includegraphics[width=\linewidth]{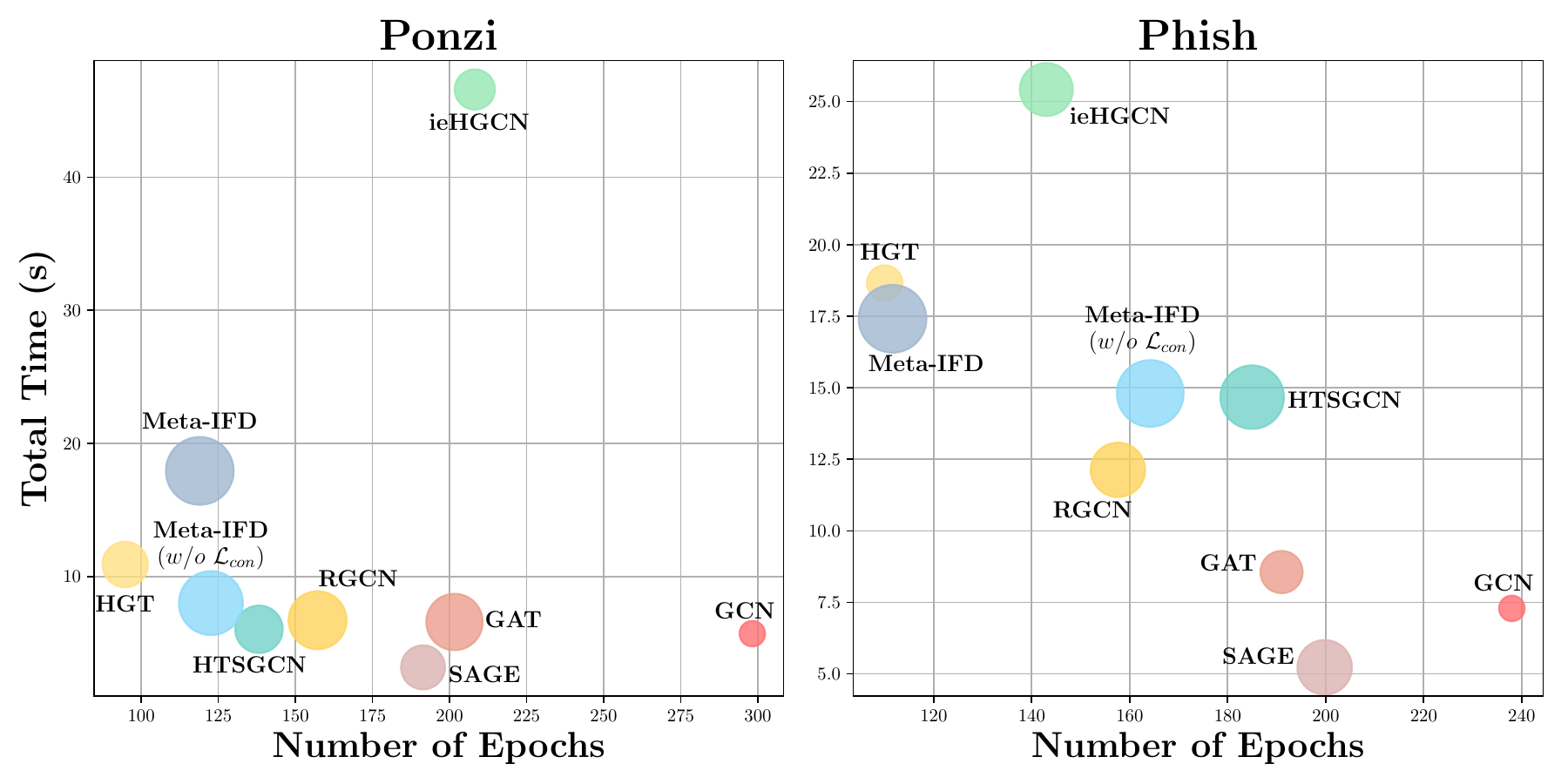}
    \caption{Performance and computational efficiency of different methods. Colors indicate different methods and sizes indicate Macro-F1 values.}
    \label{fig: efficiency}
\end{figure}

\subsection{Analysis of Computational Efficiency}
We further investigate the computational efficiency of different methods. Considering that the fine-grained interaction feature generation module can be pre-trained, we focus on evaluating the computational efficiency of the end-to-end multi-view feature learning stage. We compare only end-to-end detection methods under similar settings. Fig.~\ref{fig: efficiency} reports the total runtime and the number of epochs required for convergence under early stopping for each method. For fairness, we set the hidden layer size to 64, repeat experiments five times, and report the average results.
We can observe that while \model incurs slightly higher total runtime compared to some methods, it converges in fewer epochs and achieves superior performance. This indicates that our \model is capable of learning data features more efficiently. Additionally, when we remove the contrastive learning module—a time-intensive component—we find that the runtime decreased significantly. However, the number of epochs required for convergence increases substantially (particularly on the phishing dataset), suggesting that the self-supervised module improves the efficiency of feature learning.
The results also highlights that heterogeneous methods generally require higher time costs than homogeneous methods. This is because heterogeneous methods must differentiate between node and edge types, leading to increased computational complexity.

\section{Conclusion} \label{sec:Conclusion}
In this work, we introduce \model, a novel fraud detection framework for Ethereum that integrates fine-grained interaction feature generation, multi-view feature learning, and coarse-grained interaction feature contrast. Our experiments demonstrate that \model ~significantly outperforms existing methods in detecting Ponzi schemes and phishing scams. \model ~offers a substantial advancement in Ethereum fraud detection and provides a foundation for more secure and trustworthy blockchain ecosystems. Future work will refine and extend \model ~to address emerging fraud patterns and more blockchain platforms.

\bibliographystyle{IEEEtran} 
\bibliography{mybib,IEEEabrv}

\begin{IEEEbiography}[{\includegraphics[width=1in,height=1.25in,clip,keepaspectratio]{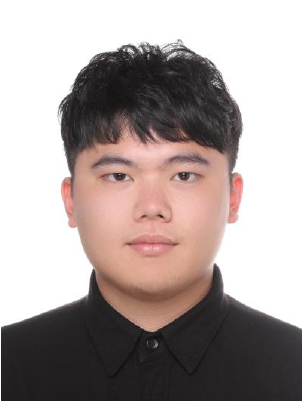}}]{Chengxiang Jin}
	received the BS degree in electrical engineering and automation from Zhejiang University of Science and Technology, Hangzhou, China, in 2021. He is currently pursuing the MS degree in control theory and engineering at Zhejiang University of Technology, Hangzhou, China. His current research interests include graph data mining and blockchain data analytics, especially for heterogeneous graph mining in Ethereum.
\end{IEEEbiography}

\begin{IEEEbiography}[{\includegraphics[width=1in,height=1.25in,clip,keepaspectratio]{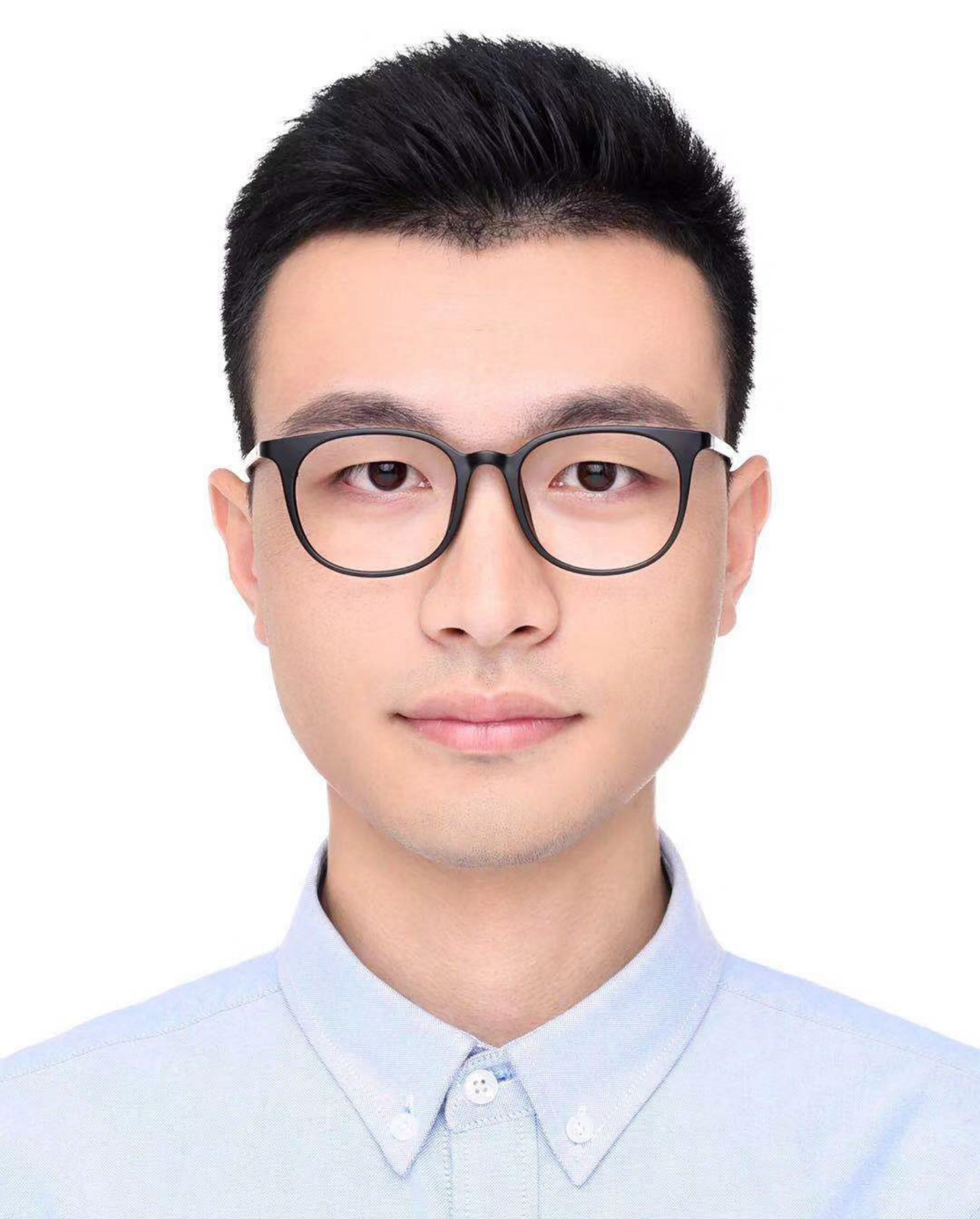}}]{Jiajun Zhou}
	received the Ph.D degree in control theory and engineering from Zhejiang University of Technology, Hangzhou, China, in 2023. He is currently a Postdoctoral Research Fellow with the Institute of Cyberspace Security, Zhejiang University of Technology. His current research interests include graph data mining, cyberspace security and data management.
\end{IEEEbiography}

\begin{IEEEbiography}[{\includegraphics[width=1in,height=1.25in,clip,keepaspectratio]{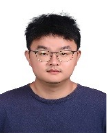}}]{Chenxuan Xie}
	received the BS degree in automation from Nanchang University, Jiangxi, China, in 2021. He is currently pursuing the MS degree in control engineering at Zhejiang University of Technology, Hangzhou, China. His current research interests include graph data mining and graph neural network, especially for graph heterophily problem and graph transformer.
\end{IEEEbiography}

\begin{IEEEbiography}[{\includegraphics[width=1in,height=1.25in,clip,keepaspectratio]{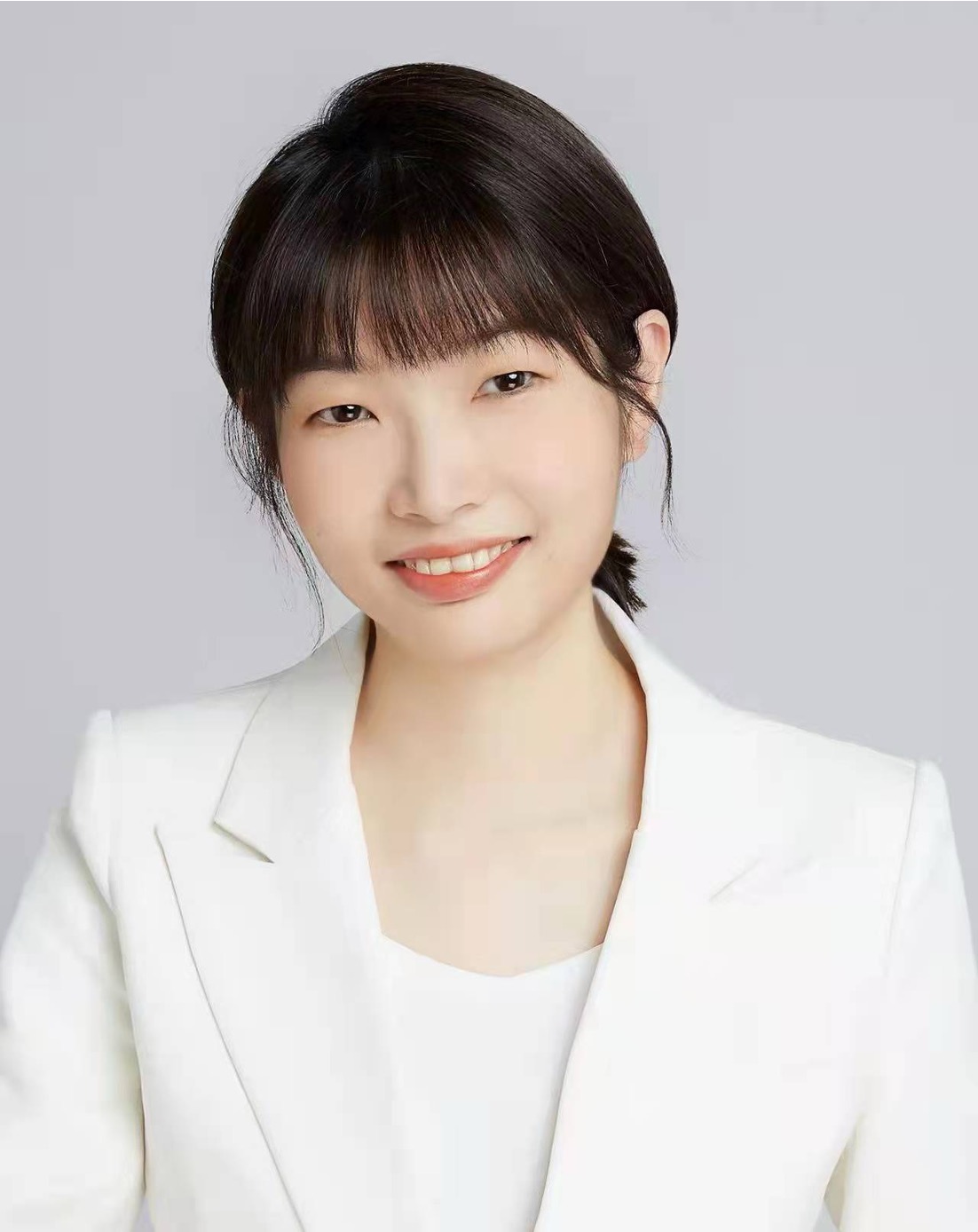}}]{Shanqing Yu}
	received the M.S. degree from the School of Computer Engineering and Science, Shanghai University, China, in 2008 and received the M.S. degree from the Graduate School of Information, Production and Systems, Waseda University, Japan, in 2008, and the Ph.D. degree, in 2011, respectively. She is currently a Lecturer at the Institute of Cyberspace Security and the College of Information Engineering, Zhejiang University of Technology, Hangzhou, China. Her research interests cover intelligent computation and data mining.
\end{IEEEbiography}

\begin{IEEEbiography}[{\includegraphics[width=1in,height=1.25in,clip,keepaspectratio]{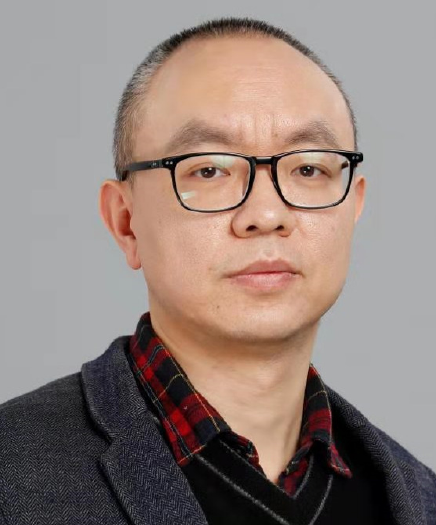}}]{Qi Xuan}(M'18) received the BS and PhD degrees in control theory and engineering from Zhejiang University, Hangzhou, China, in 2003 and 2008, respectively. He was a Post-Doctoral Researcher with the Department of Information Science and Electronic Engineering, Zhejiang University, from 2008 to 2010, respectively, and a Research Assistant with the Department of Electronic Engineering, City University of Hong Kong, Hong Kong, in 2010 and 2017. From 2012 to 2014, he was a Post-Doctoral Fellow with the Department of Computer Science, University of California at Davis, CA, USA. He is a senior member of the IEEE and is currently a Professor with the Institute of Cyberspace Security, College of Information Engineering, Zhejiang University of Technology, Hangzhou, China. His current research interests include network science, graph data mining, cyberspace security, machine learning, and computer vision.
\end{IEEEbiography}

\begin{IEEEbiography}[{\includegraphics[width=1in,height=1.25in,clip,keepaspectratio]{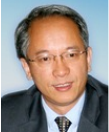}}]{Xiaoniu Yang}
  is currently a Chief Scientist with the Science and Technology on Communication
  Information Security Control Laboratory, Jiaxing, China. He is also an Academician of the Chinese Academy of Engineering and a fellow of the Chinese
  Institute of Electronics. He has published the first software radio book \emph{Software Radio Principles and Applications} [China: Publishing House of Electronics Industry, X. Yang, C. Lou, and J. Xu, 2001 (in Chinese)]. His current research interests are in software-defined satellites, big data for radio signals, and deep learning-based signal processing.
\end{IEEEbiography}

\end{document}